\documentclass[acmsmall]{acmart}
\usepackage{amsmath}
\usepackage{multirow}
\usepackage{makecell}
\usepackage{xcolor}

\AtBeginDocument{%
  }

\setcopyright{acmlicensed}
\copyrightyear{2024}
\acmYear{2024}
\acmDOI{XXXXXXX.XXXXXXX}

\acmJournal{TOMM}
\acmVolume{x}
\acmNumber{x}
\acmArticle{XXX}
\acmMonth{10}

\begin{document}

\title{Multiscale Feature Importance-based Bit Allocation for End-to-End Feature Coding for Machines}

\author{Junle Liu}
\email{liujle@mail2.sysu.edu.cn}
\affiliation{%
  \institution{School of Electronics and Communication Engineering, Sun Yat-Sen University}
  \city{Shenzhen}
  \state{Guangdong}
  \country{China}
}

\author{Yun Zhang}
\email{zhangyun2@mail.sysu.edu.cn}
\affiliation{%
  \institution{School of Electronics and Communication Engineering,  Sun Yat-Sen University}
  \city{Shenzhen}
  \state{Guangdong}
  \country{China}
}

\author{Zixi Guo}
\email{guozx29@mail2.sysu.edu.cn}
\affiliation{%
  \institution{School of Electronics and Communication Engineering,  Sun Yat-Sen University}
  \city{Shenzhen}
  \state{Guangdong}
  \country{China}
}

\author{Xiaoxia Huang}
\email{huangxiaoxia@mail.sysu.edu.cn}
\affiliation{%
  \institution{School of Electronics and Communication Engineering,  Sun Yat-Sen University}
  \city{Shenzhen}
  \state{Guangdong}
  \country{China}
}
\author{Gangyi Jiang}
\email{jianggangyi@nbu.edu.cn}
\affiliation{%
  \institution{Faculty of Information and Science and Engineering,  Ningbo University}
  \city{Ningbo}
  \state{Zhejiang}
  \country{China}
}
\begin{abstract}
\let\defaultcolor\currentcolor
Feature Coding for Machines (FCM) aims to compress intermediate features effectively for remote intelligent analytics, which is crucial for future intelligent visual applications. In this paper, we propose a Multiscale Feature Importance-based Bit Allocation (MFIBA) for end-to-end  {FCM}.  {First}, we find that the importance of features for machine vision tasks varies with the scales, object size, and image instances. Based on this finding, we propose a Multiscale Feature Importance Prediction  {(MFIP) module} to predict the importance weight for  {each scale of} features. Secondly, we propose a task loss-rate model  {to establish} the relationship between the task accuracy losses of  {using compressed features and the bitrate of encoding these features.} Finally, we develop  {a MFIBA} for end-to-end  {FCM}, which  {is able to} assign coding bits of  {multiscale} features more reasonably based on their importance. Experimental results demonstrate that when  {combined with a retained Efficient Learned Image Compression (ELIC)}, the proposed  {MFIBA} achieves an average of 38.202$\%$ bitrate savings in object detection compared to  {the anchor ELIC}.
 {Moreover, the proposed MFIBA achieves an average of 17.212$\%$ and 36.492$\%$ feature bitrate savings for instance segmentation and keypoint detection, respectively. When the proposed MFIBA is applied to the LIC-TCM, it achieves an average of 18.103$\%$, 19.866$\%$ and 19.597$\%$ bit rate savings on three machine vision tasks, respectively, which validates the proposed MFIBA has good generalizability and adaptability to different machine vision tasks and FCM base codecs.} 
\end{abstract}



\maketitle

\section{Introduction}
In recent years, the technical advancement and wide applications of multimedia have led to exponential growth in the volume of images and videos. With the rapid development of artificial intelligence, the processing of images and videos is shifting from viewed and hand-crafted with manpower to automatically analyzed by deep machine vision models. In the future, a significant portion of image and video content will be automatically processed by intelligent machine vision models as end users, rather than being viewed by humans.

 {Conventional image and video compression methods were primarily designed for human vision by exploiting signal and visual redundancies, such as JPEG and BPG for image, and High Efficiency Video Coding (HEVC) and Versatile Video Coding (VVC) \cite{vvc} for video. These methods primarily focus on human visual mechanisms while neglecting machine visual mechanisms. Consequently, the performance of the downstream machine tasks can hardly be maintained by simply using compressed images / videos from human vision oriented image / video codecs}. To adapt future intelligent visual analytics and applications, expert groups proposed new standards such as Image/Video Coding for Machines (ICM/VCM) \cite{vcm} and Joint Photographic Experts Group with Artificial Intelligence (JPEG-AI) \cite{jpegai}, which have attracted significant attention \cite{cdvampeg}. As highlighted in \cite{vcmaparadigm}, \cite{Gao2021RecentSD} and \cite{zhangyun2024}, the goals of image compression for human vision differ substantially from those for machine vision models. These initiatives \cite{overviewcdvs} emphasized the importance of addressing machine vision-specific compression, which will become an essential component of future codecs.  {In Compress-Then-Analyze (CTA) framework \cite{vcmaparadigm}, compressed images shall be reconstructed and then input for analysis. It is simple and compatible with existing image communication system, but it costs decoding overhead at the client. In Analyze-Then-Compress (ATC) paradigm \cite{vcmaparadigm}, only analytical results are required to be compressed and transmitted, which are able to achieve an extremely high compression ratio. However, ATC schemes are usually task-specific and have low generalizability or efficiency to unknown machine tasks.}

 {A number of ICM/VCM work have been developed based on conventional image/video coding framework}. Chamain \emph{et al.} \cite{endtoendICM2021} connected a machine vision task network to the decoder and trained it jointly with the codec during training. Yang \emph{et al.} \cite{towardsSIC2021} proposed to transmit the contour and color components of face images separately for face recognition tasks. Wang \emph{et al.} \cite{deepicm2023wang} optimized the object detection network to adapt to images with different losses at various bit rates. Wang \emph{et al.} \cite{endtoend2021wang} proposed an end-to-end compression with offline optimization using the Lagrange multiplier in the loss function for machine vision. Le \emph{et al.} \cite{ICMcontent2021} proposed a machine vision task network trained on uncompressed images to finetune the jointly trained machine vision task network and codec. In addition, optimization methods for bit rate allocation and quantization maps were explored. Huang \emph{et al.} \cite{analysisRD2021} proposed a Region-of-Interest for Machines (ROIM) based bit allocation optimization algorithm for object detection. It first detected image regions sensitive to the detection task using an object detection model and then allocated more bit streams to these sensitive regions. Choi \emph{et al.} \cite{TAQNforJPEG} trained a network to generate an image quantization allocation map and then performed JPEG compression on the quantized image. Cui \emph{et al.} \cite{cui2024} proposed an image compression sensing coding framework using local structural sampling to reconstruct images.  {However, these optimizations were performed based on traditional image/video codecs that were originally designed for human vision, which may degrade machine vision accuracy.}

 {To improve ICM/VCM coding efficiency, adaptations to machine vision tasks were performed on Learned Image Compression (LIC) models \cite{He2022ELICEL,LICTCM} for high compression efficiency.} Shindo \emph{et al.} \cite{shindo2024image} used regions of interest for machine vision as masks to assist the image encoding. Zhang \emph{et al.} \cite{zhang2023rethinking} decomposed images into semantic, structural, and signal layers to achieve variable-length encoding for machine vision. Chen \emph{et al.} \cite{Chen2023TransTICTT} employed prompt finetuning to adapt end-to-end image compression into machine vision-oriented codecs. Although these methods improved machine vision tasks for traditional image encoders, they still require reconstructing the image at decoder side, which not only introduces additional computational overhead but also needs to transmit an additional bit cost of image reconstruction information that may not be required in machine vision tasks. 
 {However, compressing and reconstructing signal level images brings computational overhead. In addition, the compression ratio could be improved as machine vision properties were not fully exploited.}

 {Another ICM/VCM paradigm is encoding the features extracted from images and directly performing the machine vision tasks based on the decoded features, which is newly called Feature Coding for Machines (FCM)}. Alvar \emph{et al.} \cite{multitaskcompressfeatures2019} proposed a multitask joint training method for task networks and codecs. Hyomin \emph{et al.} \cite{scalableicm_choi} \cite{Choi2021LatentSpaceSF} split the features, with one part used for machine vision tasks and the others used as enhanced features for image reconstruction. Zhang \emph{et al.} \cite{zhang2024} proposed a similar unified and scalable image compression framework for humans and machines which jointly optimizes the network in end-to-end  {manner}. Tu \emph{et al.} \cite{Tu2024} proposed a FCM by transferring knowledge from the pixel domain to the compressed domain to improve compression performance. Özyılkan \emph{et al.} \cite{latentSICM} extracted intermediate deep features from original images and used them to generatie task-specific features during feature division. In \cite{improvingmvtincompressdomain}, after extracting features from the image, a gate module was used on the encoding side to filter task-specific features. After compression, the back-end network adapted and transformed these features to perform tasks. Wang \emph{et al.} \cite{interactionorientedicm_iot} extracted features relied upon machine vision tasks from images, and the encoding side preliminarily reconstructed a rough image based on these features, transmitted the image residuals, then simultaneously performed machine vision tasks and image reconstruction at the decoding side. In \cite{Bai2021TowardsEI}, the encoder extracted and compressed image features, while the decoding side used a Transformer-based task network and an image reconstruction network. Wang \emph{et al.} \cite{multigranularity2021wangshurun} proposed a teacher-student model which used a low bit-rate decoder to assist the high bit-rate decoder in decoding, achieving multi-granularity feature compression. Codevilla \emph{et al.} \cite{Codevilla2021LearnedIC} converted the encoded latent space representations into task-adapted features through super-resolution and feature transformation modules. In \cite{Omni-ICM}, self-supervised learning was adopted to extract general features, which were then encoded, transmitted, and converted into features suitable for various machine vision tasks. Hu \emph{et al.} \cite{hu2020sensitivity} proposed a bit allocation method that utilized the sensitivity of feature channels, named the Sensitivity-Aware Bit Allocation (SABA) method. Yang \emph{et al.} \cite{compactVCM2024} experimentally validated that compressing features were more effective than compressing images.  {These methods compress and reconstruct features directly, avoiding the distortion caused by the image-level reconstruction. The FCM paradigm is superior to direct ICM. From an information theory perspective, the FCM paradigm first extracts features from images and then leverages the prior knowledge of the feature extraction model to eliminate redundant information which is irrelevant to machine vision tasks. Additionally, FCM has advantages of information encryption and low complexity by removing image reconstruction.}

 {Complex deep vision task models generally require multiscale features in the decision-making process. In this case, compressing these multiscale features becomes indispensable for remote intelligent visual analytics.} 
Liu \emph{et al.} \cite{Liu2021SemanticstoSignalSI} developed a scalable feature coding model by gradually coding and transmitting features from shallower networks, which have lower-level features and more complete information. Ning \emph{et al.} \cite{SSICMfeature} hierarchically extracted semantic features from fine-grained to coarse-grained, where the encoded features were gradually enhanced. The features for encoding and transmission were selected based on the specified target tasks in remote applications. However, these approaches could not comprehensively plan the overall features to achieve the best performance. In addition, it may include redundancies or features that are irrelevant to the task when transmitting enhanced features. In \cite{mutualFCMV2023}, only part of the features of the pyramid structure was compressed, and all scales of features were reconstructed using the transmitted partial-scale features. However, the predicted features might introduce errors due to the inaccuracy of the predictive model and cumulative feature errors in transmitted features after encoding and decoding, which can affect the performance of the machine vision model.  {In multiscale FCM, feature attributes and their dependences on machine vision tasks could be further exploited for better generalizability. The importance differences of the multiscale features and inter-scale correlations could be further exploited for higher efficiency. }

 {In this paper, we propose a Multiscale Feature Importance-based Bit Allocation (MFIBA) for end-to-end FCM, which significantly improves feature coding efficiency while maintaining the accuracies of multiple machine tasks.} The key contributions are
\begin{itemize}
\item We experimentally analyze the visual importance of  {multiscale} features in machine vision tasks and propose a  {Multiscale Feature Importance Prediction (MFIP)} to predict the  {importance of each scale of features}.
\item We propose a task loss-rate model for multiscale feature, which accurately models the relationship between task loss from compression and feature coding bits.
\item We propose a  {Multiscale Feature Importance-based Bit Allocation (MFIBA) for FCM} by exploiting the feature importance differences among scales, which reduces the feature coding bits while maintaining the accuracies of three machine vision tasks.
\end{itemize}

The paper is organized as follows: Section \uppercase\expandafter{\romannumeral2} formulates the problem of  {FCM} and statistically analyzes feature importance. Section \uppercase\expandafter{\romannumeral3} presents the proposed  {MFIBA} methods for  {FCM}. Section \uppercase\expandafter{\romannumeral4} presents the experimental results and analysis. Section \uppercase\expandafter{\romannumeral5} concludes the paper.

\section{Problem Formulation and Statistical Analysis}
In remote intelligent visual applications, the features of edge visual sensors are required to be coded and transmitted to remote data centers for visual analytics. The performance of machine vision tasks rely on  {the quality of the image feature.}  {FCM} is to code the feature effectively while maintaining the feature quality. The  {FCM} objective is
\begin{equation}
{min} \ D(\mathbf{F},\mathbf{F^{'})},\quad {s.t.}  \ R(\mathbf{F})<R,
\end{equation}
where $\mathbf{F}$ and $\mathbf{F^{'}}$ represent the original features and reconstructed features from decoding, $D(\mathbf{F}, \mathbf{F^{'}})$ is a distortion metric that measures the features quality difference and the impacts to machine vision tasks, $R(\mathbf{F})$ denotes the coding bit rate for the features $\mathbf{F}$.

Many deep networks of machine vision recommended by the VCM standard involve  {multiscale} features,  {such as} Faster Region-based Convolutional Network (R-CNN) \cite{fasterrcnn} and RetinaNet \cite{lin2018focal}  {for object detection}, Keypoint R-CNN \cite{ding2020local}  {for keypoint detection}, and Mask R-CNN \cite{maskrcnn} for  {semantic segmentation and instance segmentation}. However, features at different scales are not  {important equivalently}. 
To  {verify} this assumption, we added distortions to the  {multiscale} features (scale 0,1,...,n,P) and analyzed their impacts on object detection. We used the ResNet50 Feature Pyramid Network (FPN) \cite{7780459} as backbone to extract features at five different scales. By applying various levels of compression noise from ELIC \cite{He2022ELICEL} to each scale separately and categorizing the detection targets into three categories (large, medium, and small) following the  {definition of COCO2017 dataset}, we analyzed the relationship between features at different scales and target sizes on the COCO2017 validation dataset for the object detection.

\begin{figure*}[htb]
\begin{minipage}[b]{0.32\linewidth}
  \centering
  \centerline{\includegraphics[width=4.5cm]{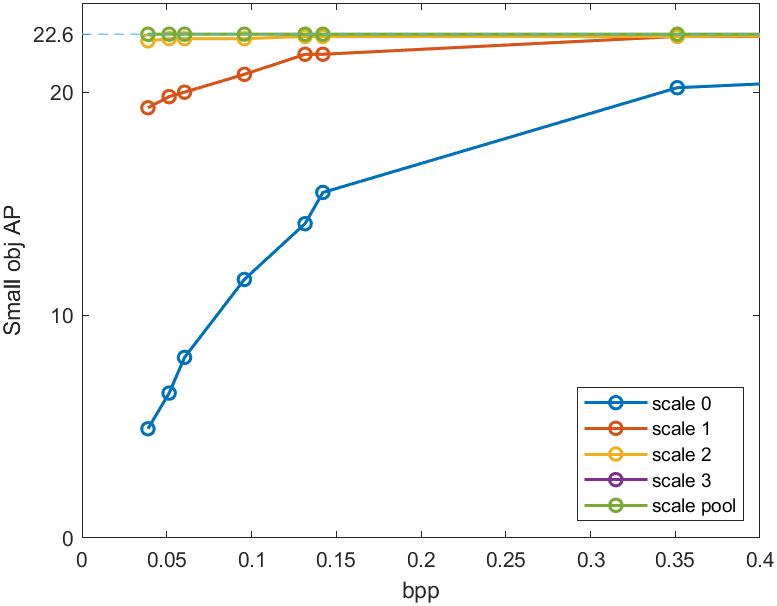}}
  \centerline{(a) Small obj. AP}\medskip
\end{minipage}
\hfill
\begin{minipage}[b]{0.32\linewidth}
  \centering
  \centerline{\includegraphics[width=4.5cm]{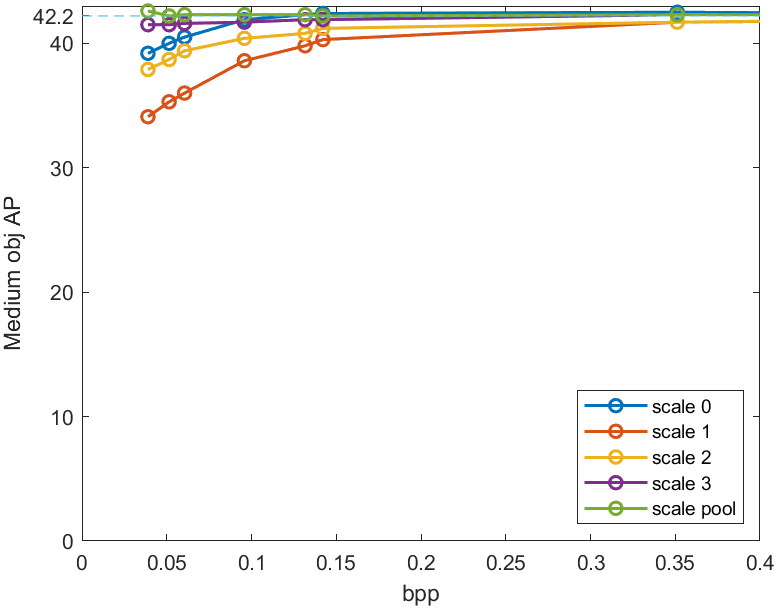}}
  \centerline{(b) Medium obj. AP}\medskip
\end{minipage}
\begin{minipage}[b]{0.32\linewidth}
  \centering
  \centerline{\includegraphics[width=4.5cm]{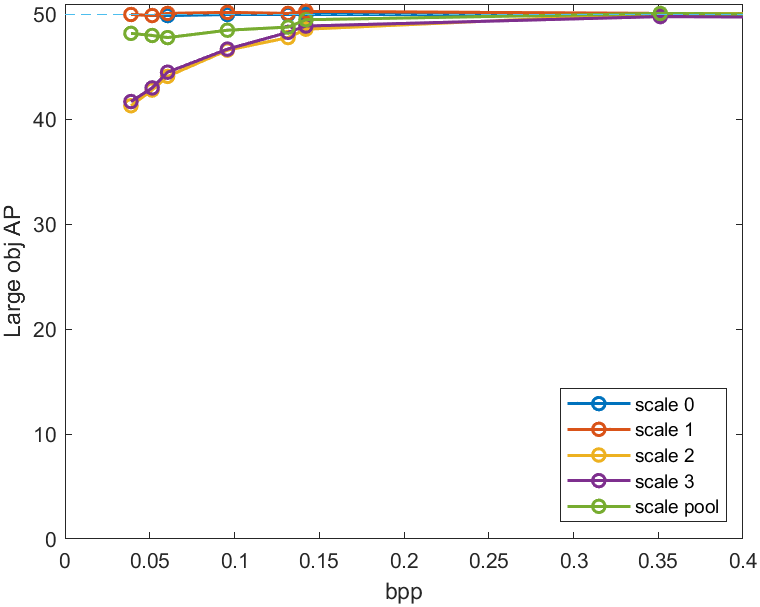}}
  \centerline{(c) Large obj. AP}\medskip
\end{minipage}
\caption{Object detection accuracies of using compressed features at different scales (following COCO2017 dataset definition) and bit rates. (a) Small objects (b) Medium objects (c) Large objects.}
\label{fig:objsize}
\end{figure*}

Fig.~\ref{fig:objsize} shows object detection results of replacing one scale of feature with the coded one at different bit rates, where the horizontal axis represents the bit rate, and the vertical axis represents the accuracy of the object detection. As shown in Fig.~\ref{fig:objsize}, the accuracies increase as the feature are compressed with higher bit rate. However, the accuracies vary with the scales and object sizes. From Fig.~\ref{fig:objsize}(a), the accuracy degrades significantly while compressing features at scale 0 for small objects, while the accuracies degrade more significantly for scale 1 and scale 3 for medium and large objects. 
It is found that small objects are more sensitive to the compression distortion of larger-scale features, such as scale 0, compared to smaller-scale features. In contrast, large objects are more sensitive to the distortion of smaller-scale features, such as scale 3 and scale pool. This is because feature receptive fields vary at different scales, larger-scale features have smaller receptive fields, making them capable of detecting small objects, but less effective at discovering large objects. Smaller-scale features aggregate information from larger receptive fields, making them capable of detecting large objects, but less sensitive to small objects. Therefore, different sizes of objects  {depend} on features from different scales. Overall, it is found that the feature importance varies with the scales and target object sizes, which shall be exploited in  {FCM}.

To achieve better machine vision tasks under bit rate constraints, it is essential to allocate more bits to the more important features. Thus, we shall assign reasonable weights to features of varying scales, indicating that their impact on the task differs after compression losses. The objective of  {MFIBA} is formulated as
\begin{equation}
{min} \ \sum\limits_{i}{w_{\emph{i}}L(\mathbf{F}_{\emph{i}},\mathbf{F}_{\emph{i}}^{'})}, \quad 
{s.t.} \ R(\mathbf{F}_{\emph{i}}) < R_T,
\label{eq:obj2}
\end{equation}
where $\emph{i} \in \left\{ 0,1,...,n,P \right\}$ denotes the $\emph{i}$-th scaled feature, $n$ is the number of scales for features, $P$ is a pooling feature from the $n$ scale of features. $w_{\emph{i}}$ is the weight assigned to each scaled feature $\emph{i}$, which depends on the instance image and task. We predict these weights in the proposed  {MFIBA} model. $\mathbf{F}_{\emph{i}}$ and $\mathbf{F}_{\emph{i}}^{'}$ are the original features and compressed features, $L(\mathbf{F}_{\emph{i}},\mathbf{F}_{\emph{i}}^{'})$ denotes the distortion loss after feature compression, $R(\mathbf{F}_{\emph{i}})$ is the coding bit rate of $\mathbf{F}_{\emph{i}}$, $R_T$ is a target bit rate.

\begin{figure}[h]
  \centering
  \centerline{\includegraphics[width=\linewidth]{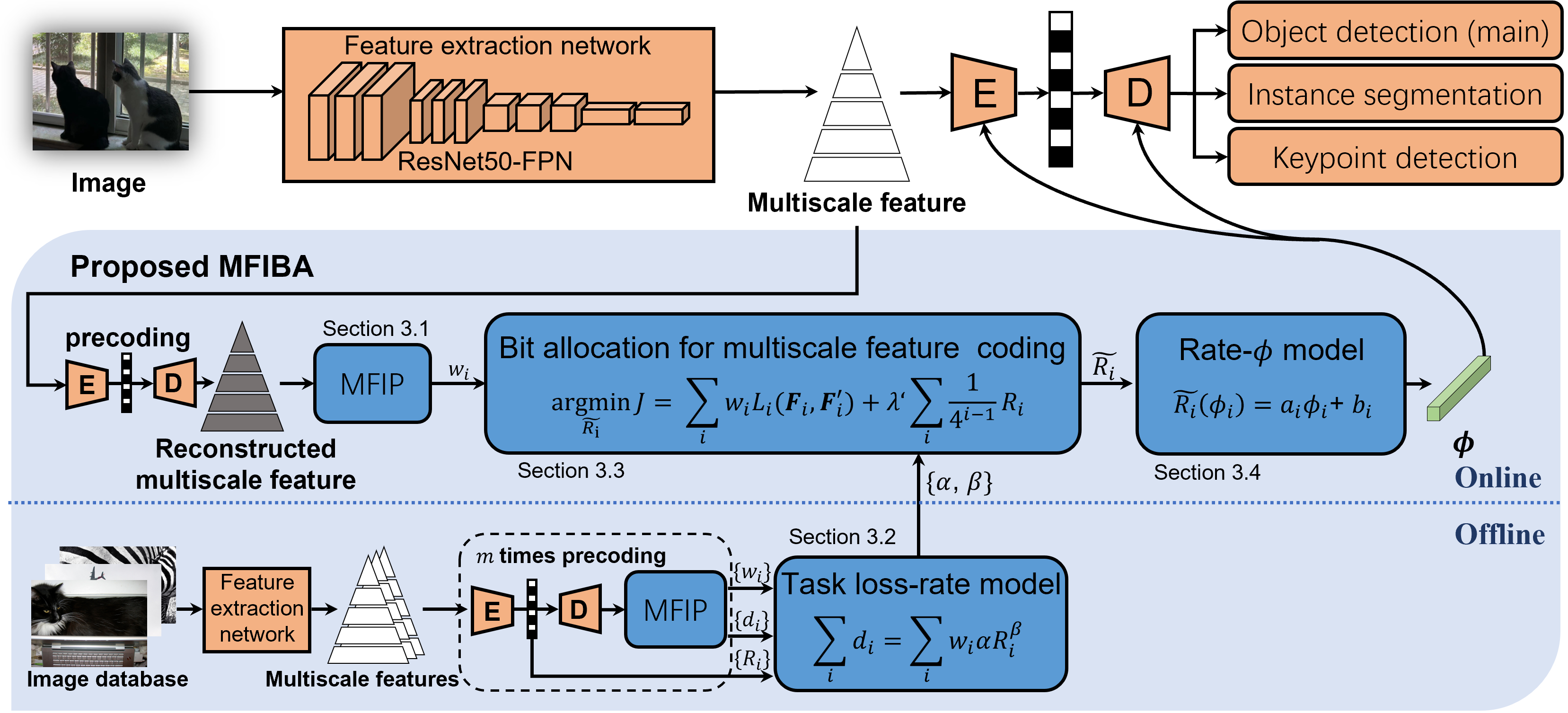}}
  \caption{ {Framework of the proposed MFIBA for FCM.}}
  \label{fig:pipline}
\end{figure}

\section{Proposed  {MFIBA} for  {FCM}}
To compress the  {multiscale} features more effectively, we propose a  {MFIBA for FCM} by considering the  {importance differences of multiscale features}.  {Fig.~\ref{fig:pipline} shows the framework of proposed MFIBA for FCM}. A feature extraction network is used to extract deep features for machine vision tasks. In this paper, we select the commonly employed ResNet50-FPN \cite{7780459} as the feature extraction network for tasks, including object detection, keypoint detection, and instance segmentation. While doing the machine vision tasks, all scales of features participate in the decision-making, which requires to code and transmit features of all scales to remote intelligent data processing center.

 {
To improve the compression efficiency of FCM, we proposed MFIBA to assign the feature coding bits more reasonably based on the importance of the feature, which includes offline and online phases. In offline phase, multiscale feature maps extracted from the image database. Then, feature importance weights $\{w_i\}$ and task losses $\{d_i\}$ from Multiscale Feature Importance Predcition (MFIP) and bit rates $\{R_i\}$ from codec multi-pass coding are input to establish a task loss-rate model. Then, in the online phase, the multiscale features are first precoded and importance weights are predicted through the MFIP. Then, based on online $\{w_i\}$ and offline task loss-rate model, bit allocation algorithm is developed to determine the budget bit rate for each scale of feature. Finally, rate-$\boldsymbol{\phi}$ model is used to map the rate to the hyperparameter of learned feature codec for coding control.}


\subsection{ {MFIP Module}}
\label{Sec:prediction}

\begin{figure}[ht]
  \centering
  \centerline{\includegraphics[width=0.85\linewidth]{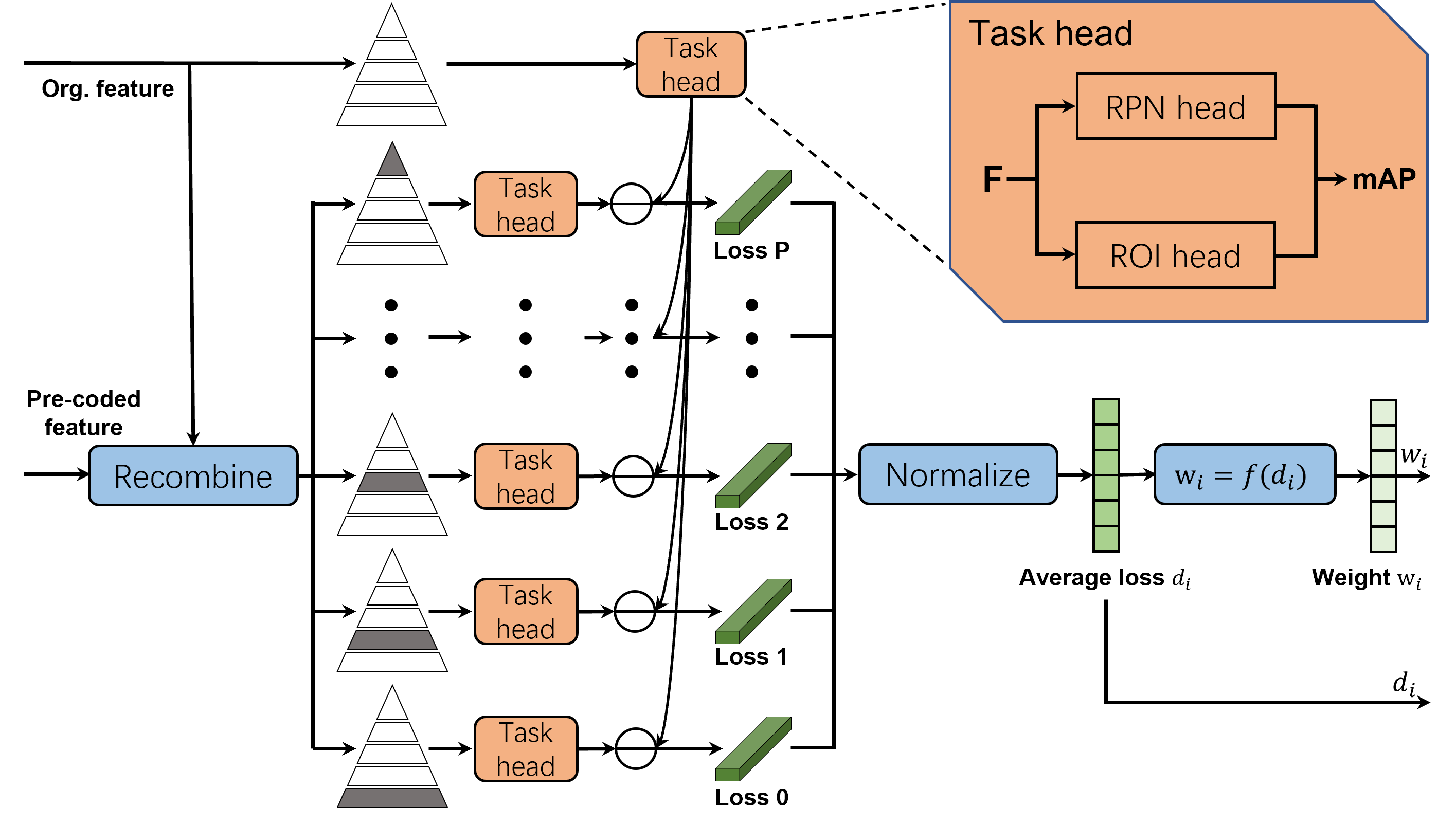}}
  \caption{Flowchart of the proposed  {MFIP module}.}
  \label{fig:weight_model}
\end{figure}

Based on the observations in Section \uppercase\expandafter{\romannumeral2},  {multiscale} feature shall be given different importance. To measure and predict the importance of each feature scale, we propose  {MFIP module} for each instance image, as shown in Fig.~\ref{fig:weight_model}. First, we precode  {multiscale} features at different quality levels. Then, the  {MFIP module} combines the  {precoded} results with the original features, replacing some scales of the original features with the  {precoded} features. The recombined features are put in the task heads \cite{fasterrcnn} to obtain detection results of different scales and distortions. These results are compared with the detection results of using lossless  {multiscale} features to derive loss vectors of each feature scale at different quality levels. Considering the different dynamic ranges of these losses, these loss vectors are normalized. Then, the task loss for each scale feature at different qualities is averaged to get the average loss $d_\emph{i}$. Finally, the importance weights of  {multiscale} features are predicted from the average of normalized task losses.

\begin{figure}[htb]
\begin{minipage}[b]{0.5\linewidth}
  \centering
  \centerline{\includegraphics[width=\linewidth]{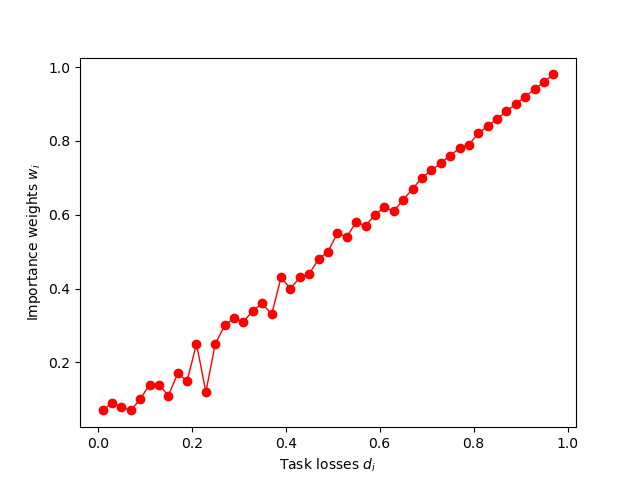}}
  \caption{ {Relationship between task loss $d_\emph{i}$ and importance weight $w_\emph{i}$ across all images.}}
  \label{w_d}
\end{minipage}
\hfill
\begin{minipage}[b]{0.40\linewidth}
  \centering
  \centerline{\includegraphics[width=\linewidth]{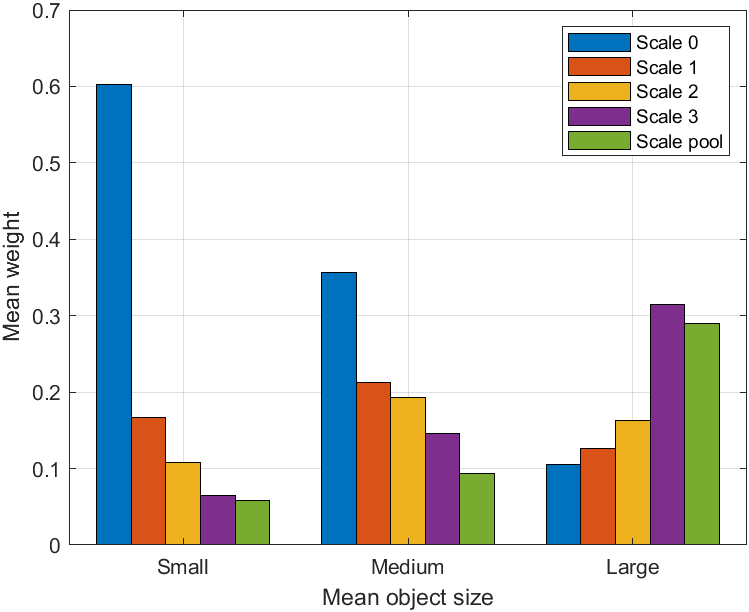}}
  \caption{Relationship between the object target sizes and the average weight $\hat{w_\emph{i}}$.}
  \label{w_result}
\end{minipage}
  
\end{figure}

Since  {object detection} task loss is the final metric to  {optimize}, we believe that there is a high correlation between the importance weights of the features and the average loss values in Fig.\ref{fig:weight_model}. To analyze this correlation, the weights can be expressed as $\hat{w_\emph{i}}=f(d_\emph{i})$, where $d_\emph{i}$ is the task accuracy loss of $\emph{i}$-th scale feature, $\hat{w_\emph{i}}$ is the predicted weight and $f()$ is the mapping function.  {To investigate the relationship between weights $w_\emph{i}$ and task losses $d_\emph{i}$, we visualize their correlation across images from the COCO 2017 validation dataset in Fig.~\ref{w_d}. We can observe that there is a high linear correlation between them.} For simplicity, we predict $\hat{w_\emph{i}}$ directly with $d_\emph{i}$,  {which is} expressed as $\hat{w_\emph{i}}=d_\emph{i}$. 

If more computational resources are avaiable, we  {can employ} a line search method to find the optimal $w_\emph{i}$.  {Since loss $d_\emph{i}$ is continuous, it is challenging to find the optimal $w_\emph{i}$ directly. The search mainly  {includes} four key steps: 1) we first discretize $d_\emph{i}$ as an initial prediction of the weights $\hat{w_\emph{i}}$. 2) We apply the line search method starting from the initial $\hat{w_\emph{i}}$. Specifically, for each scale feature, we quantize $\hat{w_\emph{i}}$ to two decimals and perform the rate allocation in the MFIBA using two calibrated weights, i.e., $\hat{w_\emph{i}}+0.01$ and $\hat{w_\emph{i}}-0.01$. 3) The compression efficiency is evaluated based on the object detection accuracy from compressed features and coding bit rate. If better compression efficiency is achieved by using either $\hat{w_\emph{i}}+0.01$ or $\hat{w_\emph{i}}-0.01$. Then, an additional step,  $\hat{w_\emph{i}}+0.02$ or $\hat{w_\emph{i}}-0.02$ will be searched and compared. 4) Steps 2)-3) are repeated until the optimal importance weight for each scale is found. Therefore, the MFIBA algorithm can allocate bitrates more effectively to different scale features, achieving better compression performance on object detection.}

 {Furthermore}, we analyzed the relationship between the feature weights $\hat{w_\emph{i}}$ calculated by the  {MFIP module} and different object sizes in  {the} images. The statistical results are shown in Fig.~\ref{w_result}. It can be observed that in images with smaller average target sizes, the average weight of larger-scale features is higher, while in images with larger average target sizes, the weights are more inclined towards smaller-scale features. Thus, the  {MFIP module} effectively allocates feature weights based on target sizes.  {Notably, the inherent similarities in the underlying mechanisms of machine vision tasks suggest that importance weights can be generalized to tasks beyond object detection \cite{Gao2021RecentSD}.}

\subsection{Task Loss-Rate Model}

 {To model the relationship between task loss and bit rate, we propose a feature importance weighted task loss-rate model based on the Cauchy-distributed Rate-Distortion (RD) model \cite{RDO2015}.} Due to the varying importance of  {multiscale} features in machine vision tasks, we predicted the importance of each scale of features $w_i$ using the  {MFIP module}. Then, the machine vision task loss-rate model is constructed from the predicted feature weights and the feature distortions, which can be modeled by the exponential model \cite{RDO2015}. Therefore, the task loss-rate model can be presented as
\begin{equation}
{
  \sum\limits_{i}{d_{\emph{i}}} = ~{\sum\limits_{i}{{w_{\emph{i}}}~\alpha{R_{\emph{i}}}^{- \beta}}},
\label{Eq:LRmodel}
}
\end{equation}
where $R_\emph{i}$ is the bit rate for compressing features at the ${i}^{th}$ scale, $w_{\emph{i}}$ is the weights of  {multiscale} features,  {$d_\emph{i}$ is the task loss obtained from the MFIP module, $\alpha$ and $\beta$ are factors for the Cauchy-distributed loss-rate model, which are derived from offline multi-pass precoding of feature dataset.} 

\begin{figure}[htb]
\begin{minipage}[b]{0.48\linewidth}
  \centering
  \centerline{\includegraphics[width=0.8\linewidth]{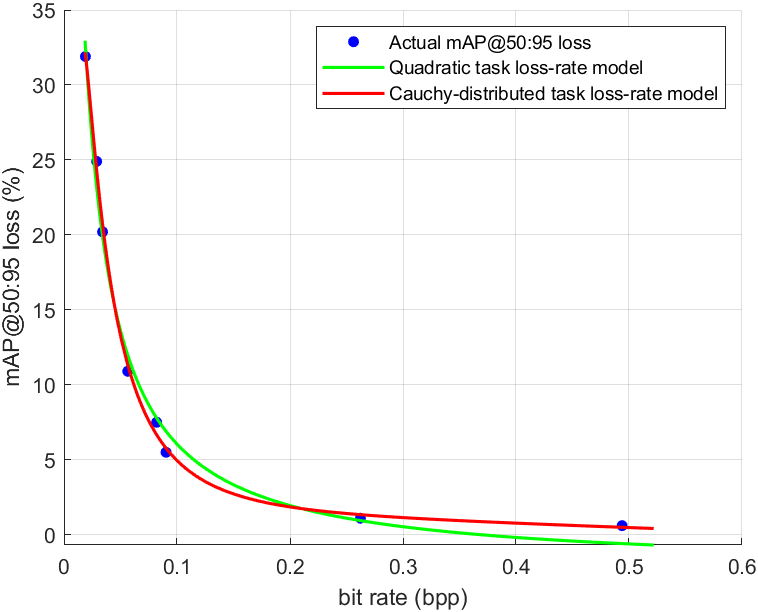}}
  \centerline{(a)}\medskip
\end{minipage}
\hfill
\begin{minipage}[b]{0.48\linewidth}
  \centering
  \centerline{\includegraphics[width=0.8\linewidth]{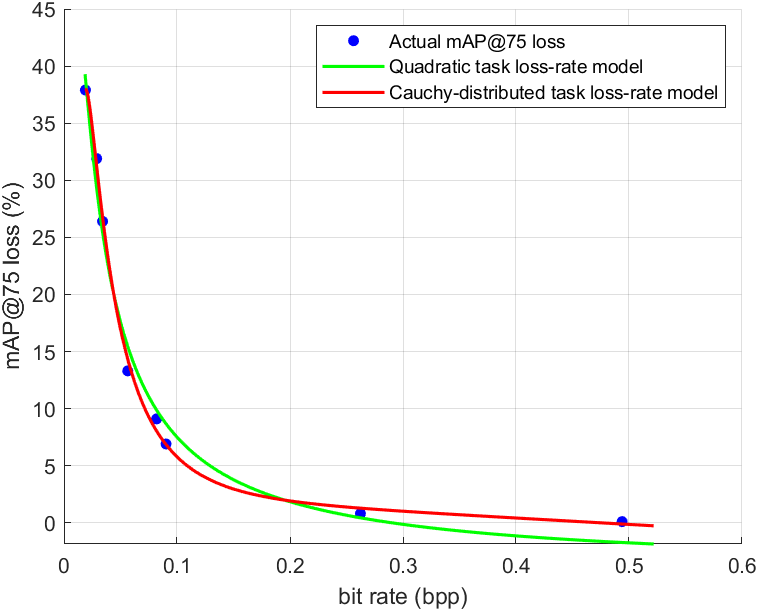}}
  \centerline{(b)}\medskip
\end{minipage}
  \caption{ {Relationship between task loss and bit rate, (a) mAP@50:95, (b) mAP@75.}}
  \label{loss-rate}
\end{figure}

 {To evaluate the accuracy of Cauchy-distributed loss-rate model, we performed coding experiments for multiscale features of all images in the COCO2017 validation dataset, where used as the base codec. Object detection using Faster R-CNN was performed by using the reconstructed features from different coding bit rates. The detection accuracy was measured with mAP@50:95 and mAP@75. Fig. \ref{loss-rate} presents the fitting results of the Cauchy-distributed task loss-rate model in Eq. \ref{Eq:LRmodel} and quadratic task loss-rate model \cite{Quadratic} for comparison, where y-axis represents loss in object detection accuracy on mAP@50:95 and mAP@75, and x-axis represents bit rate in bpp.} We can observe that the model fits the real data  {better}. The fitting accuracies of the models are illustrated in Table \ref{tab:loss-rate}. The Correlate Coefficients (CC) of the cauchy-distributed task loss-rate model  {in mAP@50:95 and mAP@75} are 0.9991 and 0.9989, respectively,  {which are higher than those of quadratic model.} The Root Mean Squared Error (RMSE)  {in mAP@50:95 and mAP@75} are 0.4551 and 0.6217, respectively,  {which are lower than those of the quadratic task loss-rate model. It shows that the Cauchy-distributed task loss-rate model can accurately model the relationship between task loss and rate. In addition, the accurate loss-rate model will consequently improve the bit allocation accuracy in MFIBA.}

\begin{table}
  \caption{ {CC and RMSE of the task loss-rate models, where the best are in bold.}}
  \label{tab:loss-rate}
  \begin{center}
  \begin{tabular}{c c c c}
    \hline
    Task loss-rate models & Evaluation metrics & mAP@50:95 & mAP@75\\
    \hline
    \multirow{2}{*}{Quadratic} & CC & 0.9956 & 0.9930\\
    & RMSE & 1.0078 & 1.5966\\
    \multirow{2}{*}{Cauchy-distributed} & CC & \textbf{0.9991} & \textbf{0.9989}\\
    & RMSE & \textbf{0.4551} & \textbf{0.6217}\\
    \hline
\end{tabular}
\end{center}
\end{table}

\subsection{ {Bit Allocation for Multiscale Feature Coding}}
\label{bitallocation}
Based on the predicted feature importance weights $\hat{w_i}$ and the introduced Lagrange Multiplier $\lambda$, the objective function in Eq. \ref{eq:obj2} can be updated by minimizing a loss function $J$ as
\begin{equation}
  J = ~{\sum\limits_{i}{\hat{w}_{\emph{i}}L_{\emph{i}}(\mathbf{F}_{\emph{i}},\mathbf{F}_{\emph{i}}^{'})}} + \lambda{\sum\limits_{i} R_{\emph{i}}},
\end{equation}
where $L_\emph{i}(\mathbf{F}_{\emph{i}},\mathbf{F}_{\emph{i}}^{'})$ is the distortion loss of the $\emph{i}$-th scale feature, $R_\emph{i}$ is the bit rate of encoding $\emph{i}$-th scale feature. $\emph{i} \in \left\{ 0,1,...,n,P \right\}$ denotes the $\emph{i}$-th scale of feature, $n$ is the number of feature scales and $P$ denotes the pooling feature. Note that for low complexity, $\hat{w_i}$ is a predicted version of ${w_i}$ by using the  {MFIP module} in Section \ref{Sec:prediction}.
Considering the different sizes of feature maps across various scales, let $S_0$ be the size of the $0$-th feature map, since the compression rate $R_{\emph{i}}$ is based on the size of each scale of feature, the bit rate of the overall objective function is based on the sum of bits across all scales of features, the  {final} optimization objective is updated as
\begin{equation}
     {
    J = {\sum\limits_{i}{\hat{w}_{\emph{i}}L_{\emph{i}}(\mathbf{F}_{\emph{i}},\mathbf{F}_{\emph{i}}^{'})}} + \lambda '{\sum\limits_{i}{\frac{1}{k^{\emph{i}}}R_\emph{i}}},}
\label{eq:objectfun1}
\end{equation}
where  {$\lambda^{'} = \lambda S_{0}$}, $k$ is a scaling ratio of features between scales $i$ and $i+1$, $k$ is 4 for ResNet50-FPN.

Applying Eq. \ref{Eq:LRmodel} to Eq. \ref{eq:objectfun1}, it is easy to prove that Eq. \ref{eq:objectfun1} is convex. So we can get the optimal solution by calculating the partial derivative of each variable with respect to the optimization function. By setting the partial derivatives to zero, the optimal budget rate {$\widetilde{R_\emph{i}}$} for $i^{th}$ scale feature can be determined as
\begin{equation}
     {\widetilde{R_\emph{i}} = \left(\frac{\hat{w_{\emph{i}}}\alpha\beta}{\lambda^{'}/k^{i}} \right)^{\frac{1}{\beta + 1}}}, 
\end{equation}
where $\emph{i} \in \left\{ 0,1,...,n,P \right\}$, $\hat{w_\emph{i}}$ is the predicted weights obtained from  {MFIP module. Based on this solution, it is found that the bit budget $\widetilde{R_\emph{i}}$ of $i^{th}$ scale feature increases as the feature importance $\hat{w_\emph{i}}$ increases, and vice versa}.

\subsection{ {Rate-$\boldsymbol{\phi}$ Model for End-to-end Feature Coding}}
Since the feature codecs were based on the end-to-end image codecs, the coding bit rate was controlled by the hyperparameter $\phi$ within the compression network.  {In end-to-end LIC network, the loss function is conventionally formulated as $L = R + \phi D $, where $\phi$ serves as the Lagrange multiplier balancing the trade-off between the distortion loss $D$ and the rate $R$. In end-to-end LIC codecs, the bit budget $\widetilde{R_{\emph{i}}}$ cannot be applied to the end-to-end LIC codecs for coding control directly, but to be converted to the multiplier $\phi$. To this end, Rate-$\boldsymbol{\phi}$ model is developed to establishes an explicit mapping between $\boldsymbol{\phi}$ and $R_{\emph{i}}$,} $\boldsymbol{\phi} = \{ \phi_\emph{0}, \phi_\emph{1}, ..., \phi_\emph{n}, \phi_\emph{P} \}$,  {and $R_{\emph{i}}$ is the coding rate of different feature scales.} In this work, we model the relationship betwen the bit rate $R_i$ and $\phi_\emph{i}$ with a linear function, which can be expressed as
\begin{equation}
  R_{\emph{i}}(\phi_\emph{i}) = a_{\emph{i}}\times \phi_\emph{i} + b_{\emph{i}},
\label{Eq:linear}
\end{equation}
where $a_{\emph{i}}$ and $b_{\emph{i}}$ are weighting factors of scale $i$.

\begin{figure}[htb]
  \centering
  \centerline{\includegraphics[width=0.5\linewidth]{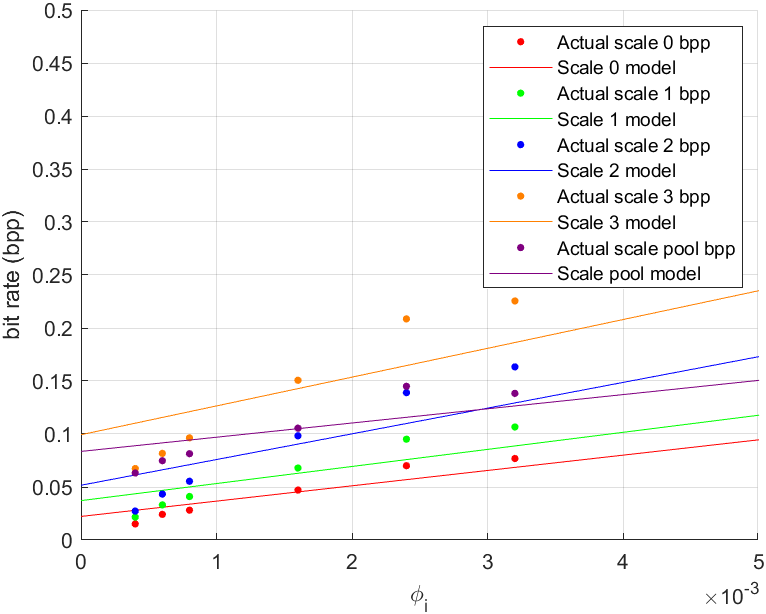}}
  \caption{Relationship between the bit rate $R_{\emph{i}}(\phi_\emph{i})$ and $\phi_\emph{i}$.}.
  \label{r-q}
\end{figure}

To testify the accuracy of Eq. \ref{Eq:linear}, we experimentally analyzed the $R_{\emph{i}}(\phi_\emph{i})$ function while coding each scale of feature. Fig.~\ref{r-q} shows the relationship between $R_{\emph{i}}(\phi_\emph{i})$ and $\phi_\emph{i}$ from feature coding, where dots are from real coded features from COCO2017 and the lines are fitted curves using Eq. \ref{Eq:linear}. The lines generally fit well with the collected data for  {multiscale} features. Table ~\ref{tab:r-q} accuracies in terms of CC and RMSE of fitted results by using Eq. \ref{Eq:linear}, where P stands for the pooling feature from  {multiscale} features. The CC between the actual coding bit rate and the $R_{\emph{i}}(\phi_\emph{i})$ model of the five scaled features are 0.9938, 0.9855, 0.9823, 0.9774, and 0.9768, respectively. The RMSE of them are 0.0084, 0.0145, 0.0238, 0.0321, and 0.0181, respectively. It can be observed that the $R_{\emph{i}}(\phi_\emph{i})$ function effectively and accurately captured the relationship between the hyperparameter $\boldsymbol{\phi}$ and the resulting coding bit rate, demonstrating its ability to precisely model how changes in $\boldsymbol{\phi}$ influence compression dynamics.

\begin{table}
  \caption{CC and RMSE of the $R_{\emph{i}}(\phi_\emph{i})$ model.}
  \label{tab:r-q}
  \begin{center}
  \begin{tabular}{c c c c c c}
    \hline
    Feature scale & 0 & 1 & 2 & 3 & P\\
    \hline
    CC & 0.9938 & 0.9855 & 0.9823 & 0.9774 & 0.9768\\
    RMSE & 0.0084 & 0.0145 & 0.0238 & 0.0321 & 0.0181\\
    \hline
\end{tabular}
\end{center}
\end{table}

 {Finally, based on feature bit budget $\widetilde{R_\emph{i}}$ and Eq. \ref{Eq:linear}, hyperparameter $\phi_\emph{i}$ for $i^{th}$ feature coding control is obtained as
\begin{equation}
{
  \phi_\emph{i} = (\widetilde{R_{\emph{i}}} - b_{\emph{i}})/a_{\emph{i}}.
\label{Eq:final phi}
}
\end{equation}
By implementing Eq. \ref{Eq:final phi}, the MFIBA enables a more efficient feature bit allocation for FCM.}

\section{Experimental Results and Analysis}
\subsection{Experimental Settings}
We selected the object detection,  {instance segmentation and keypoint detection} as the machine vision tasks, which utilize the  {multiscale} features in  {decision-making}. We randomly selected 2,000 images from the COCO2017 validation dataset to evaluate the compression network. Additionally, considering that the data distribution between natural image datasets and feature map datasets differs, future codecs targeting machine vision may rely on models trained on feature map datasets rather than natural images. We constructed a feature map dataset by randomly selecting 40,000 images from the COCO2017 training dataset. We used the ResNet50-FPN \cite{7780459} backbone to extract 5-scale features, resulting in 200,000 feature matrices. Each matrix, with 256 channels, was split into 3-channel patches, producing 200,000 3-channel images for retraining the end-to-end feature coding models. 

We chose Faster R-CNN \cite{fasterrcnn} as the target network  {for object detection}, which is  {recommended} by the VCM standards group. This network uses ResNet50 \cite{7780459} backbone extract features at five scales, which collectively contribute to task decision-making.  {Additionally, to validate the generalizability of the proposed MFIBA on other machine tasks, instance segmentation with Mask R-CNN \cite{maskrcnn} and keypoint detection with Keypoint R-CNN \cite{ding2020local} were used to evaluate the quality of compressed features, respectively.} We selected the Effcient Learned Image Compression (ELIC) \cite{He2022ELICEL} and  as the basic feature codec, on which we applied the proposed  {MFIBA} to improve the feature coding efficiency. We utilized 8 bit rate points from pre-training to establish task loss-rate model. " {MFIBA+ELIC}" denotes that the proposed MFIBA is applied to ELIC, while "SABA+ELIC" denotes that the SABA \cite{hu2020sensitivity} is applied to ELIC. "Fineture MFIBA" indicates that the importance weights of feautures are further caliberated with line search. In addition, since the ELIC was initially proposed for human vision oriented image coding, we retrained ELIC on the feature dataset, which is denoted as "Retrained ELIC". Correspondingly,  " {MFIBA+Retrained ELIC}" and " {Finetuned MFIBA+Retrained ELIC}" are MFIBA applied to the  {retrained} ELIC and  {finetuned} MFIBA applied to the  {retrained} ELIC. VTM 17.0 is a test model for VVC \cite{vvc}. To further verify the generalizability of the proposed  {MFIBA} to different codecs, we also applied the  {MFIBA} with the same parameters to another advanced Learned Image Compression Transformer-CNN Mixture (LIC-TCM)\cite{LICTCM},labeled as "LIC-TCM".

We used mean average precision with Intersection over Union (IoU) thresholds from 50$\%$ to 95$\%$ (i.e., mAP@50:95) and 75$\%$ (i.e., mAP@75) as evaluation metrics to measure the accuracy of three  {machine vision tasks}, including object detection, instance segmentation and keypoint detection. The feature bit rate is measured with bits per pixel (bpp). The Bjøntegaard-Delta Bit Rate (BDBR) was calculated by comparing with the anchor ELIC \cite{He2022ELICEL} and LIC-TCM \cite{LICTCM}.

\begin{table}
  \caption{BDBR of the proposed  {MFIBA+ELIC} and benchmark schems compared to ELIC  {on object detection. [Unit:$\%$ ].}}
  \label{result tabel 1}
  \begin{center}
  \begin{tabular}{c c c}
    \hline
    Coding Schemes & mAP@50:95 & mAP@75\\
    \hline
     {SABA \cite{hu2020sensitivity}+ELIC} & 2.113 & 6.073\\
    VTM 17.0 \cite{vvc} & -5.661 & -18.625\\
     {Retrained ELIC} & -13.477 & -14.179\\
     {MFIBA+ELIC} & -16.118 & -21.586\\
     {Finetuned MFIBA+ELIC} & -16.834 & -21.922\\
     {MFIBA+Retrained ELIC} & -30.373 & -32.304\\
     {Finetuned MFIBA+Retrained ELIC} & \textbf{-35.992} & \textbf{-38.202}\\
    \hline
\end{tabular}
\end{center}
\end{table}

\begin{figure*}[htb]
\begin{minipage}[b]{0.48\linewidth}
  \centering
  \centerline{\includegraphics[width=0.8\linewidth]{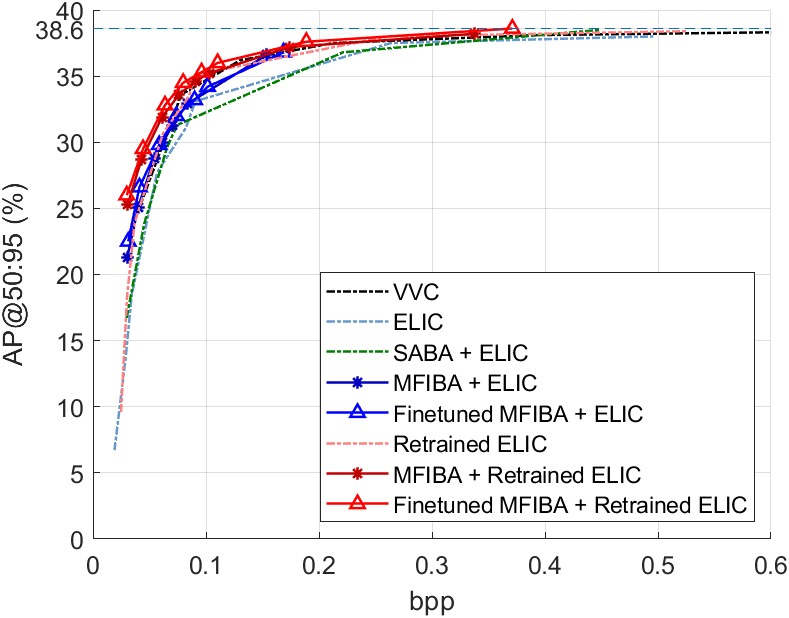}}
  \centerline{(a)}\medskip
\end{minipage}
\hfill
\begin{minipage}[b]{0.48\linewidth}
  \centering
  \centerline{\includegraphics[width=0.8\linewidth]{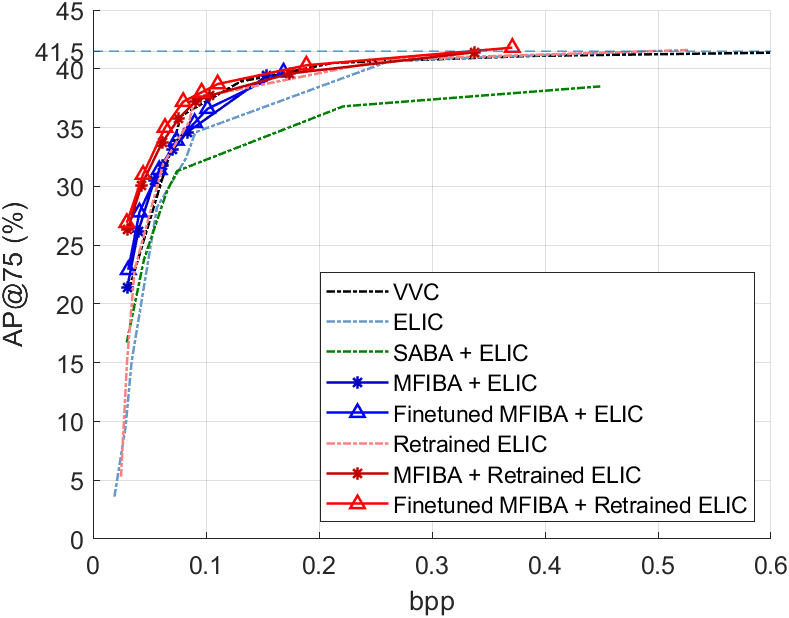}}
  \centerline{(b)}\medskip
\end{minipage}
\caption{Coding performance of the Proposed  {MFIBA based} on ELIC  {for object detection.} (a) mAP@50:95 (b) mAP@75.}
\label{fig:result1}
\end{figure*}

\begin{figure*}[htb]

\begin{minipage}[b]{.32\linewidth}
  \centering
  \centerline{\includegraphics[width=4.0cm]{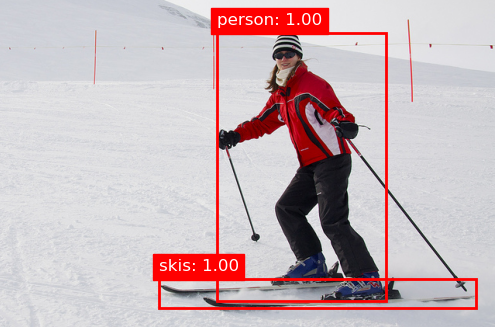}}
  \centerline{(a) Ground truth}\medskip
\end{minipage}
\hfill
\begin{minipage}[b]{.32\linewidth}
  \centering
  \centerline{\includegraphics[width=4.0cm]{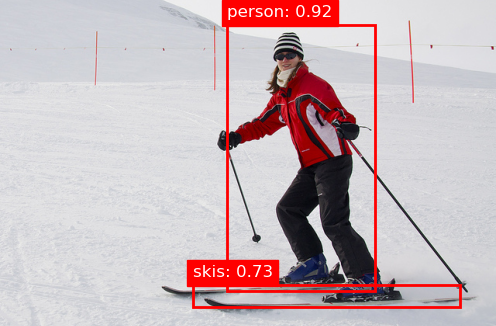}}
  \centerline{(b) ELIC (0.172 bpp)}\medskip
\end{minipage}
\hfill
\begin{minipage}[b]{.32\linewidth}
  \centering
  \centerline{\includegraphics[width=4.0cm]{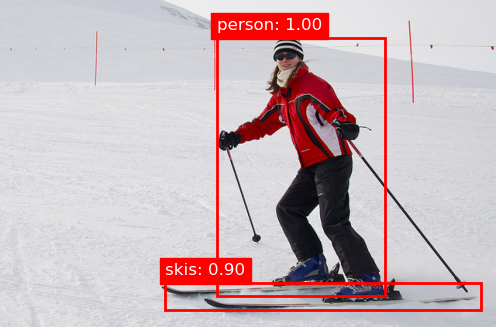}}
  \centerline{(c)  {MFIBA+ELIC} (0.176 bpp)}\medskip
\end{minipage}

\begin{minipage}[b]{0.32\linewidth}
  \centering
  \centerline{\includegraphics[width=4.0cm]{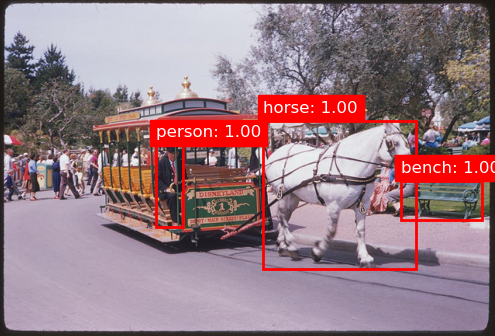}}
  \centerline{(d) Ground truth}\medskip
\end{minipage}
\hfill
\begin{minipage}[b]{0.32\linewidth}
  \centering
  \centerline{\includegraphics[width=4.0cm]{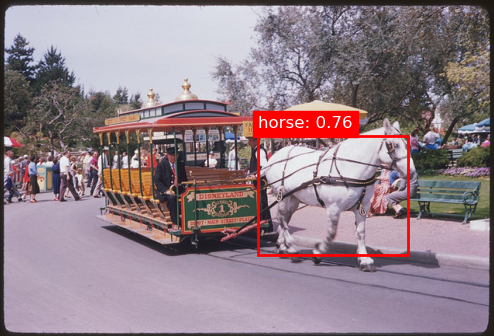}}
  \centerline{(e) ELIC (0.237 bpp)}\medskip
\end{minipage}
\hfill
\begin{minipage}[b]{0.32\linewidth}
  \centering
  \centerline{\includegraphics[width=4.0cm]{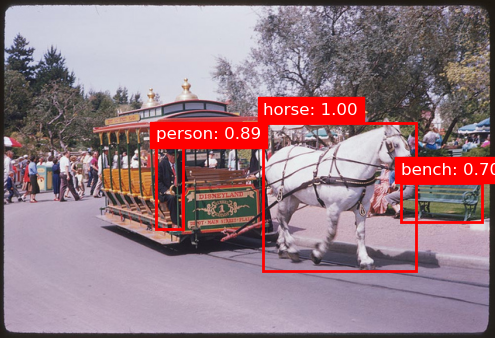}}
  \centerline{(f)  {MFIBA+ELIC} (0.252 bpp) }\medskip
\end{minipage}

\caption{ {Visual comparison of object detection using compressed features from different codecs.} (a) and (d): Ground truth, (b) and (e): ELIC, (c) and (f):  The Proposed  {MFIBA+ELIC}, (a)-(c): image of person and skis, (d)-(f) image of a horse, a person and a bench.}
\label{visualization}
\end{figure*}
\subsection{Compression Performance and Analysis based on ELIC}

Table \ref{result tabel 1} presents quantitative BDBR of the proposed  {MFIBA} on ELIC and benchmark schems as comparing with the ELIC  {on object detection.} The  {SABA+ELIC} is inferior to pre-trained ELIC with bit rates increasing of 2.113$\%$ under mAP@50:95 and 6.073$\%$ under the mAP@75 on average. The VTM outperforms the ELIC with 5.661$\%$ and 18.625$\%$ BDBR gains when task performance is measured with mAP@50:95 and mAP@75, respectively. If ELIC is retained with feature set, 13.477$\%$ and 14.179$\%$ bit rat saving can be achieved under mAP@50:95 and mAP@75, respectively. It indicates that re-training using feature dataset  to transfer LIC of natural images to FCM is somewhat effective. The proposed  {MFIBA+ELIC} achieved bit rate savings of 16.118$\%$ under the comprehensive evaluation metric mAP@50:95 and 21.586$\%$ under mAP@75. When sufficient computational resources were available, the  {Finetuned MFIBA+ELIC} further improved compression efficiency, and achieved bit rate savings of 21.922$\%$ bit rate under mAP@75 and 16.834$\%$ under mAP@50:95. Based on the retrained ELIC,  {MFIBA+Retrained ELIC} achieved bit rate savings of 30.373$\%$ under mAP@50:95 and 32.304$\%$ under mAP@75, while the algorithm using the finetuned weights saved bit rate up to 35.992$\%$ and 38.202$\%$ of the bit rate. Note that the weights of  {MFIBA} used here were from the pre-trained ELIC and were not recalibrated. Fig.~\ref{fig:result1} illustrates the bit rate allocation results of the ELIC codec, both pre-trained and trained on the feature map training set and the comparison method  {SABA+ELIC} \cite{hu2020sensitivity} and VTM. It can be observed that the proposed  {Finetuned MFIBA+Retrained ELIC} achieves the best performance and the  {MFIBA+ELIC} significantly outperforms the  {SABA+ELIC}. 

These coding results indicate that the proposed  {MFIBA} method significantly enhanced the  {FCM} performance. Also, it can effectively adapt to different learnable deep codecs. Based on the pre-trained ELIC, the  {Finetuned MFIBA+ELIC} slightly outperformed the  {MFIBA+ELIC}. When utilizing the retrained codec, finetuned parameters exhibit stronger generalizability across the retrained encoder compared to  {MFIBA} without finetuning, as the finetuned parameters align more closely with the real importance weights of  {multiscale} features.

In addition to BDBR, Fig.~\ref{visualization} presents the object detection results of two images from COCO2017 dataset whose features were extracted by the original source image, or encoded by ELIC, and  {MFIBA} optimized ELIC and then detected. We can observe from Fig.~\ref{visualization}.(b) that the ELIC achieves 0.92 and 0.73 confidence for the person and skis when the coding bit is 0.172 bpp. For the proposed  {MFIBA} optimized ELIC, shown as Fig.~\ref{visualization}. (c), it achieves 1.00 and 0.9 confidence, respectively, for the person and skis when the coding bit is 0.176 bpp, which is better and more consistent with the results of the ground truth. Similar results can be found for Fig.~\ref{visualization}(d)-(f). Overall,  {with} similar bit rates, the  {MFIBA} optimized ELIC can achieve higher detection accuracy, probability, and confidence compared to the ELIC.

\begin{table}
  \caption{BDBR of the proposed  {MFIBA+}LIC-TCM compared to LIC-TCM  {on object detection. [Unit:$\%$]}}
  \label{result tabel 2}
  \begin{center}
  \begin{tabular}{c c c}
    \hline
    Coding schemes & mAP@50:95 & mAP@75\\
    \hline
     {SABA \cite{hu2020sensitivity}+LIC-TCM} & -3.572 & 10.391\\
    VTM 17.0 \cite{vvc} & -11.818 & 0.261\\
     {MFIBA+LIC-TCM} & \textbf{-18.103}  & -13.783\\
     {Finetuned MFIBA+LIC-TCM} & -15.798 & \textbf{-15.374}\\
    \hline
  \end{tabular}
  \end{center}
\end{table}

\begin{figure*}[htb]
\begin{minipage}[b]{0.48\linewidth}
  \centering
  \centerline{\includegraphics[width=0.8\linewidth]{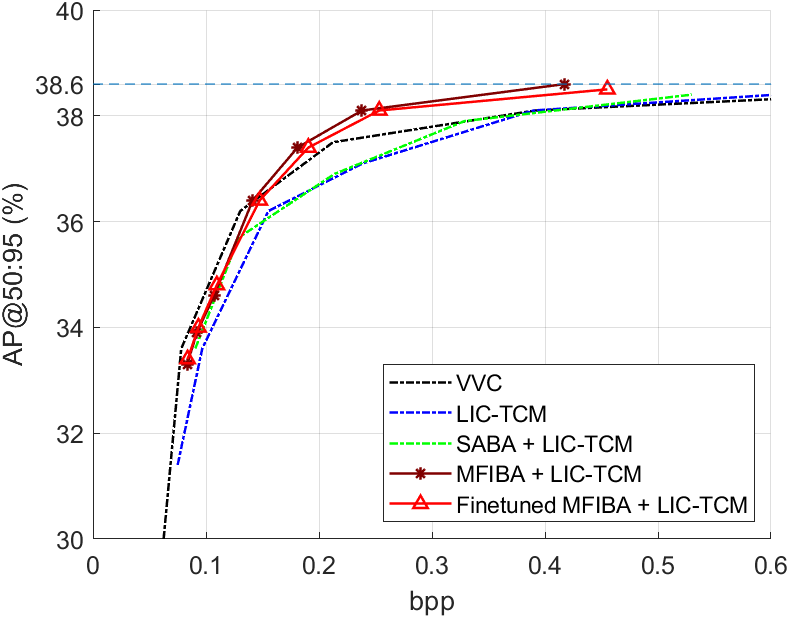}}
  \centerline{(a)}\medskip
\end{minipage}
\hfill
\begin{minipage}[b]{0.48\linewidth}
  \centering
  \centerline{\includegraphics[width=0.8\linewidth]{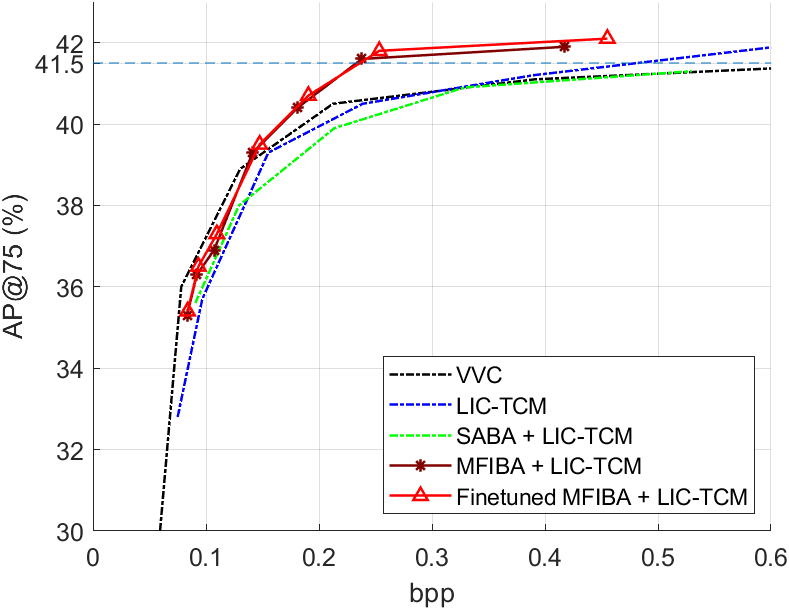}}
  \centerline{(b)}\medskip
\end{minipage}
\caption{Coding Performance of the Proposed  {MFIBA} based on LIC-TCM  {on object detection}, (a) mAP@50:95 (b) mAP@75.}
\label{fig:result2}
\end{figure*}

\subsection{Compression Performance and Analysis based on LIC-TCM}

To further validate the performance of the proposed  {MFIBA}, it was applied to another end-to-end image codec, LIC-TCM, to optimize the feature coding efficiency.  {Similarly} to the experimental settings in the previous subsection  {for the object detection}, comparative  {FCM} experiments were performed by applying the proposed  {MFIBA} to the LIC-TCM and compared with the LIC-TCM, VVC, and  {SABA+LIC-TCM}. Note that the parameters we used are calculated from the ELIC codec without updating importance weights. Table \ref{result tabel 2} presents the quantitative BD-BR comparisons while encoding the deep features, where the original LIC-TCM was used as  {a reference} in calculating BD-BR. We can observe that the  {SABA+LIC-TCM} improves compression performance by saving 3.572$\%$ bit rate under mAP@50:95 metric, while the VVC saves an average of 11.818$\%$ bit rate but is inferior to LIC-TCM by increasing 0.261$\%$ of bit rate under mAP@75.  {MFIBA+LIC-TCM} achieves an average of 18.103$\%$ bit rate savings under mAP@50:95 and 13.783$\%$ under mAP@75. The  {Finetuned MFIBA+LIC-TCM} saved up to 15.798$\%$ of the bit rate under mAP@50:95 and 15.374$\%$ under mAP@50:95.

Fig.~\ref{fig:result2} illustrates  {FCM} results of the proposed  {MFIBA} applied to LIC-TCM, compared with the benchmark coding schemes. It can be observed that the proposed  {MFIBA+LIC-TCM} significantly outperformed the  {SABA+LIC-TCM} and was comparable to VVC at low bit rates while outperforming VVC at high bit rates. The experimental results proved that the proposed  {MFIBA} has good generalizability and can be adapted to different learnable deep codecs. However, the weights of  {MFIBA} finetuned on the ELIC encoder did not improve the encoding performance further as it was applied to the LIC-TCM encoder as compared with the LIC-TCM using the direct  {MFIBA}. This is because the finetuned  {MFIBA} weights are sensitive and potentially correlated with the network structure of the deep encoder. To further improve encoding performance on LIC-TCM, updating the finetuned weights based on the LIC-TCM is preferred.  {In general}, the proposed  {MFIBA} can significantly improve the  {FCM} performance using LIC-TCM.

\begin{figure*}[htb]
\begin{minipage}[b]{0.48\linewidth}
  \centering
  \centerline{\includegraphics[width=0.8\linewidth]{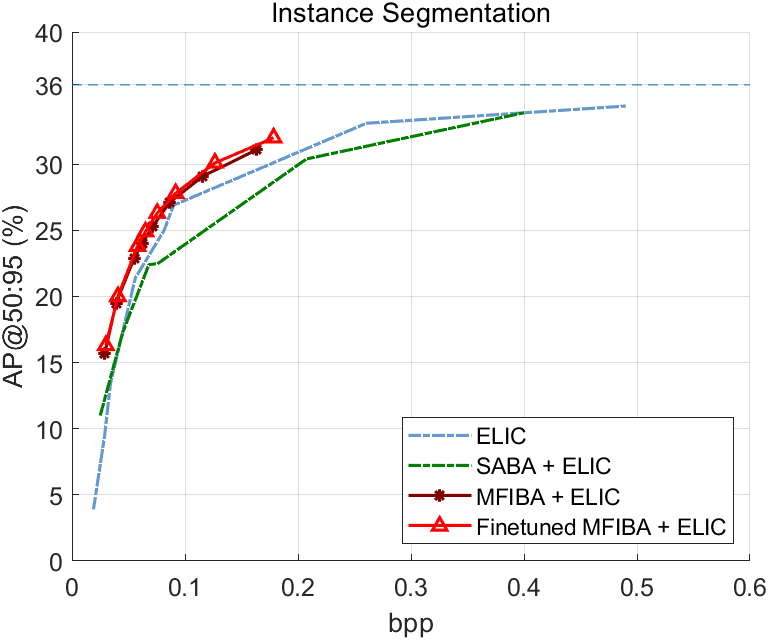}}
  \centerline{(a)}\medskip
\end{minipage}
\hfill
\begin{minipage}[b]{0.48\linewidth}
  \centering
  \centerline{\includegraphics[width=0.8\linewidth]{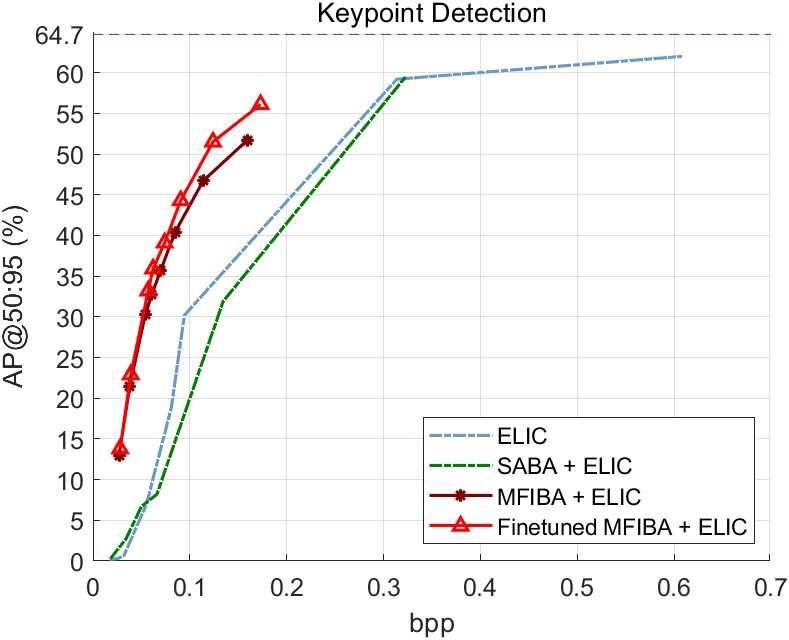}}
  \centerline{(b)}\medskip
\end{minipage}
\begin{minipage}[b]{0.48\linewidth}
  \centering
  \centerline{\includegraphics[width=0.8\linewidth]{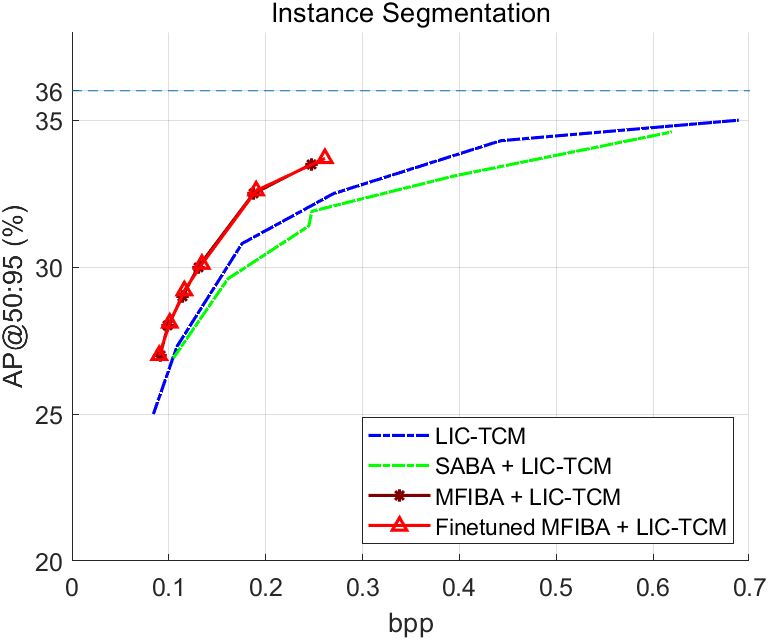}}
  \centerline{(c)}\medskip
\end{minipage}
\hfill
\begin{minipage}[b]{0.48\linewidth}
  \centering
  \centerline{\includegraphics[width=0.8\linewidth]{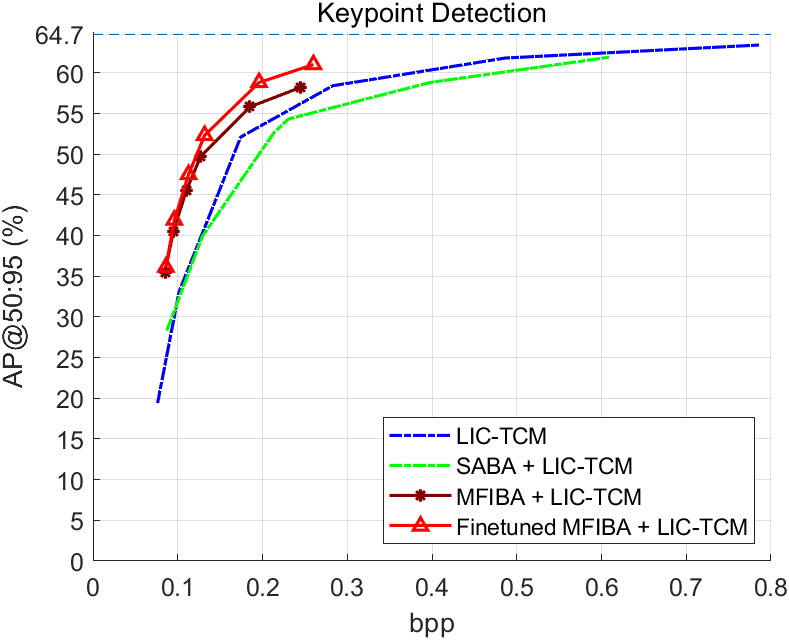}}
  \centerline{(d)}\medskip
\end{minipage}
\caption{ {RD Performance of the proposed MFIBA+ELIC/LIC-TCM on instance segmentation and keypoint detection, where the accuracies are measured with mAP@50:95. (a)(c) instance segmentation, (b)(d) keypoint detection, (a)(b) ELIC based, (c)(d)LIC-TCM based.}}
\label{fig:other tasks}
\end{figure*}

\begin{table}
  \caption{ {BDBR of the proposed MFIBA on instance segmentation and keypoint detection, where the accuracy is measured with mAP@50:95 and mAP@75. [Unit:$\%$]}}
  \label{result other tasks}
  \begin{center}
  \begin{tabular}{c c c c c c}
    \hline
    \thead{Machine Vision Task} & \thead{Coding Schemes} & \thead{mAP@50:95 \\ (+ ELIC)} & \thead{mAP@50:95 \\ (+ LIC-TCM)} & \thead{mAP@75 \\ (+ ELIC)} & \thead{mAP@75 \\ (+ LIC-TCM)}\\
    \hline
    \multirow{3}{*}{Instance Segmentation} & SABA \cite{hu2020sensitivity} & 11.793 & 13.129 & 16.141 & 14.508\\
    & MFIBA & -14.156 & -19.588 & -20.721 & -25.953\\
    & Finetuned MFIBA & \textbf{-17.212} & \textbf{-19.866} & \textbf{-23.373} & \textbf{-26.799}\\
    \hline
    \multirow{3}{*}{Keypoint Detection} & SABA \cite{hu2020sensitivity} & 8.288 & 17.523 & 0.398 & 18.921\\
    & MFIBA & -34.002 & -10.852 & \textbf{-17.029} & -4.124\\
    & Finetuned MFIBA & \textbf{-36.492} & \textbf{-19.597} & -16.775 & \textbf{-12.952}\\
    \hline
\end{tabular}
\end{center}
\end{table}

\subsection{ {Compression Performance and Analysis on Other Machine Vision Tasks}}
 {To validate that the proposed MFIBA can adapt to other machine vision tasks, we proposed coding experiments on the instance segmentation and keypoint detection based on ELIC and LIC-TCM. The compression results are shown in Fig.~\ref{fig:other tasks} and the quantitative results of BD-BR are shown in Table \ref{result other tasks}.} 
 {First, in the instance segmentation, the SABA+ELIC increases BDBR by 11.793$\%$ on average compared to the anchor ELIC. The main reason is that SABA was proposed for object detection, but is no longer effective as being applied to instance segmentation directly. Compared to ELIC, the proposed MFIBA+ELIC achieves an average of 14.156$\%$ bit rate savings, which is significant. If weights of the MFIBA are calibrated with finetuning, the finetuned MFIBA+ELIC achieves an average of 17.212$\%$ bit rate savings. If apply the MFIBA to LIC-TCM, the MFIBA and the  {Finetuned MFIBA+LIC-TCM} achieve an average of 19.588$\%$ and 19.866$\%$, respectively. }
 {Second, in the keypoint detection, SABA+ELIC increases bit rates by 8.288$\%$ compared to the anchor ELIC. The MFIBA and the finetuned MFIBA are able to improve the ELIC and achieve 34.002$\%$ and 36.492$\%$ BDBR gains on average, respectively. If apply the MFIBA to LIC-TCM, the MFIBA and the finetuned MFIBA achieve 10.852$\%$ and 19.597$\%$ BDBR gains, respectively. }
 {Third, similar results can be found when the accuracies of instance segmentation and keypoint detection are measured with mAP@75. In summary, these results show that the proposed MFIBA is generalizable to multiple machine tasks, including instance segmentation and keypoint detection.}

\subsection{ {Computational Complexity Analysis}}

 {We performed theoretical and experimental analysis on the complexity of the proposed MFIBA. Encoding and decoding experiments were carried out on an Intel Core i9-10900 CPU and a NVIDIA GeForce RTX 3090 GPU. In encoding, the time of precoding multiscale features is denoted as $t_{pre}$; the time of estimating importance weights in MFIP and performing bit allocation in the MFIBA algorithm is denoted as $t_{assign}$. The actual feature coding time is denoted as $t_{enc}$. In decoding, the decoding time is denoted as $t_{dec}$, and the time of performing task model for task performance evaluation is denoted as $t_{task}$. Let $m$ be the number of precoding with different bit rates and $n+2$ be the number of different scale features contained in the multiscale features ($\emph{i} \in {0,1,...,n,P}$), then $t_{pre} = m(t_{enc} + t_{dec}) + m(n+2)t_{task}$. The $t_{pre}$ will be more than $m$ times of $t_{enc} + t_{dec}$.}

\begin{table}
  \caption{ {Encoding and decoding time of the proposed MFIBA and benchmarks.[Unit:s]}}
  \label{tab:time}
  \begin{center}
  \begin{tabular}{c c c c c c c}
    \hline
    \multirow{2}{*}{Base codec} & \multirow{2}{*}{Methods} & \multicolumn{3}{c}{Encoding Time (s)} & \multicolumn{2}{c}{Decoding Time (s)}\\
    & & $t_{pre}$ & $t_{assign}$ & $t_{enc}$ & $t_{dec}$ & $t_{task}$\\
    \hline
    \multirow{3}{*}{ELIC} & ELIC & - & - & 2.395 & 1.598 & 0.394\\
    & SABA\cite{hu2020sensitivity}+ELIC & 6.456 & 0.001 & 2.396 & 1.596 & 0.396\\
    & MFIBA+ELIC & 47.541 & 0.001 & 2.403 & 1.596 & 0.394\\
    \hline
    \multirow{3}{*}{LIC-TCM} & LIC-TCM & - & - & 2.468 & 1.646 & 0.391\\
    & SABA\cite{hu2020sensitivity}+LIC-TCM & 6.581 & 0.001 & 2.473 & 1.651 & 0.395\\
    & MFIBA+LIC-TCM & 24.621 & 0.001 & 2.465 & 1.645 & 0.388\\
    \hline
\end{tabular}
\end{center}
\end{table}

 {Table \ref{tab:time} shows encoding and decoding time of the proposed MFIBA and the benchmarks. We have the following five observations. First, $t_{pre}$ of the SABA and MFIBA are 6.456$s$  and 47.541$s$, respectively. $t_{assign}$ of the SABA and MFIBA are 0.001$s$, which is negligible. $t_{enc}$ of ELIC, SABA+ELIC and the proposed MFIBA are 2.395$s$, 2.396$s$ and 2.403$s$, respectively, which are similar. Second, $t_{dec}$ and $t_{task}$ of the ELIC, SABA+ELIC and MFIBA+ELIC are quite similar as the decoder are generally not changed. Third, similar results can be found for LIC-TCM based coding optimization, where $t_{pre}$ of the MFIBA is 24.621$s$ and higher than that of SABA. Fourth, the $t_{pre}$ based on ELIC is higher than that of $t_{pre}$ based on LIC-TCM. The main reason is the precoding times $m$ are 8 for ELIC and 6 for LIC-TCM, respectively. Higher model accuracy could be achieved with more pre-coding times and task evaluation. Finally, the number of parameters in MFIBA+ELIC and MFIBA+LIC-TCM are 144.65MB and 534.41MB, respectively, which are the same as those of ELIC and LIC-TCM. It means MFIBA generally does not increase the size of model parameter.}

\section{Conclusions}

In this paper, we proposed a  {Multiscale} Feature Importance-based Bit Allocation ( {MFIBA}) for  {Feature Coding for Machines (FCM)}, which adaptively allocated coding bits based on the varying importance on different scales of features in machine vision tasks. Firstly, we found that the importance of feature varied with the scales, object size, and image instances and proposed an  {Multiscale Feature Importance Prediction (MFIP) module} to predict feature importance. Secondly, we proposed a task loss-rate model to build the relationship between bit rate and task losses for machine vision tasks. Finally, we developed the  {MFIBA} for end-to-end  {FCM}, where the optimal coding parameters for each scale of feature were calculated by solving a  {MFIBA} optimization function. Experiments demonstrated that the proposed  {MFIBA} can save an average of 21.922$\%$ bit rate based on Efficient Learned Image Compression (ELIC) and can achieve an average of 38.201$\%$ bit rate saving for codecs retrained on features compared to anchor ELIC.  {In addition, the MFIBA has good generalizability to different machine vision tasks and base codecs.}

\bibliographystyle{ACM-Reference-Format}
\bibliography{sample-base}


\begin{thebibliography}{45}


\ifx \showCODEN    \undefined \def \showCODEN     #1{\unskip}     \fi
\ifx \showDOI      \undefined \def \showDOI       #1{#1}\fi
\ifx \showISBNx    \undefined \def \showISBNx     #1{\unskip}     \fi
\ifx \showISBNxiii \undefined \def \showISBNxiii  #1{\unskip}     \fi
\ifx \showISSN     \undefined \def \showISSN      #1{\unskip}     \fi
\ifx \showLCCN     \undefined \def \showLCCN      #1{\unskip}     \fi
\ifx \shownote     \undefined \def \shownote      #1{#1}          \fi
\ifx \showarticletitle \undefined \def \showarticletitle #1{#1}   \fi
\ifx \showURL      \undefined \def \showURL       {\relax}        \fi
\providecommand\bibfield[2]{#2}
\providecommand\bibinfo[2]{#2}
\providecommand\natexlab[1]{#1}
\providecommand\showeprint[2][]{arXiv:#2}

\bibitem[Alvar and Bajić(2019)]%
        {multitaskcompressfeatures2019}
\bibfield{author}{\bibinfo{person}{Saeed~Ranjbar Alvar} {and} \bibinfo{person}{Ivan~V. Bajić}.} \bibinfo{year}{2019}\natexlab{}.
\newblock \showarticletitle{Multi-Task Learning with Compressible Features for Collaborative Intelligence}. In \bibinfo{booktitle}{\emph{2019 IEEE International Conference on Image Processing (ICIP)}}. \bibinfo{publisher}{IEEE}, \bibinfo{address}{Taipei, Taiwan}, \bibinfo{pages}{1705--1709}.
\newblock
\showISSN{2381-8549}
\urldef\tempurl%
\url{https://doi.org/10.1109/ICIP.2019.8803110}
\showDOI{\tempurl}


\bibitem[Ascenso et~al\mbox{.}(2023)]%
        {jpegai}
\bibfield{author}{\bibinfo{person}{João Ascenso}, \bibinfo{person}{Elena Alshina}, {and} \bibinfo{person}{Touradj Ebrahimi}.} \bibinfo{year}{2023}\natexlab{}.
\newblock \showarticletitle{The JPEG AI Standard: Providing Efficient Human and Machine Visual Data Consumption}.
\newblock \bibinfo{journal}{\emph{IEEE MultiMedia}} \bibinfo{volume}{30}, \bibinfo{number}{1} (\bibinfo{year}{2023}), \bibinfo{pages}{100--111}.
\newblock
\urldef\tempurl%
\url{https://doi.org/10.1109/MMUL.2023.3245919}
\showDOI{\tempurl}


\bibitem[Bai et~al\mbox{.}(2021)]%
        {Bai2021TowardsEI}
\bibfield{author}{\bibinfo{person}{Yuanchao Bai}, \bibinfo{person}{Xu Yang}, \bibinfo{person}{Xianming Liu}, \bibinfo{person}{Junjun Jiang}, \bibinfo{person}{Yaowei Wang}, \bibinfo{person}{Xiangyang Ji}, {and} \bibinfo{person}{Wen Gao}.} \bibinfo{year}{2021}\natexlab{}.
\newblock \showarticletitle{Towards End-to-End Image Compression and Analysis with Transformers}.
\newblock \bibinfo{journal}{\emph{ArXiv}}  \bibinfo{volume}{abs/2112.09300} (\bibinfo{year}{2021}).
\newblock
\urldef\tempurl%
\url{https://api.semanticscholar.org/CorpusID:245329388}
\showURL{%
\tempurl}


\bibitem[Bross et~al\mbox{.}(2021)]%
        {vvc}
\bibfield{author}{\bibinfo{person}{Benjamin Bross}, \bibinfo{person}{Ye-Kui Wang}, \bibinfo{person}{Yan Ye}, \bibinfo{person}{Shan Liu}, \bibinfo{person}{Jianle Chen}, \bibinfo{person}{Gary~J. Sullivan}, {and} \bibinfo{person}{Jens-Rainer Ohm}.} \bibinfo{year}{2021}\natexlab{}.
\newblock \showarticletitle{Overview of the Versatile Video Coding (VVC) Standard and its Applications}.
\newblock \bibinfo{journal}{\emph{IEEE Transactions on Circuits and Systems for Video Technology}} \bibinfo{volume}{31}, \bibinfo{number}{10} (\bibinfo{year}{2021}), \bibinfo{pages}{3736--3764}.
\newblock
\urldef\tempurl%
\url{https://doi.org/10.1109/TCSVT.2021.3101953}
\showDOI{\tempurl}


\bibitem[Chamain et~al\mbox{.}(2021)]%
        {endtoendICM2021}
\bibfield{author}{\bibinfo{person}{Lahiru~D. Chamain}, \bibinfo{person}{Fabien Racapé}, \bibinfo{person}{Jean Bégaint}, \bibinfo{person}{Akshay Pushparaja}, {and} \bibinfo{person}{Simon Feltman}.} \bibinfo{year}{2021}\natexlab{}.
\newblock \showarticletitle{End-to-End optimized image compression for machines, a study}. In \bibinfo{booktitle}{\emph{2021 Data Compression Conference (DCC)}}. \bibinfo{publisher}{IEEE}, \bibinfo{address}{Snowbird, UT, USA}, \bibinfo{pages}{163--172}.
\newblock
\showISSN{2375-0359}
\urldef\tempurl%
\url{https://doi.org/10.1109/DCC50243.2021.00024}
\showDOI{\tempurl}


\bibitem[Chen et~al\mbox{.}(2020)]%
        {Quadratic}
\bibfield{author}{\bibinfo{person}{Yi Chen}, \bibinfo{person}{Sam Kwong}, \bibinfo{person}{Mingliang Zhou}, \bibinfo{person}{Shiqi Wang}, \bibinfo{person}{Guopu Zhu}, {and} \bibinfo{person}{Yi Wang}.} \bibinfo{year}{2020}\natexlab{}.
\newblock \showarticletitle{Intra Frame Rate Control for Versatile Video Coding with Quadratic Rate-Distortion Modelling}. In \bibinfo{booktitle}{\emph{ICASSP 2020 - 2020 IEEE International Conference on Acoustics, Speech and Signal Processing (ICASSP)}}. \bibinfo{pages}{4422--4426}.
\newblock
\urldef\tempurl%
\url{https://doi.org/10.1109/ICASSP40776.2020.9054633}
\showDOI{\tempurl}


\bibitem[Chen et~al\mbox{.}(2023)]%
        {Chen2023TransTICTT}
\bibfield{author}{\bibinfo{person}{Yi-Hsin Chen}, \bibinfo{person}{Ying Weng}, \bibinfo{person}{Chia-Hao Kao}, \bibinfo{person}{Cheng Chien}, \bibinfo{person}{Wei-Chen Chiu}, {and} \bibinfo{person}{Wenmin Peng}.} \bibinfo{year}{2023}\natexlab{}.
\newblock \showarticletitle{TransTIC: Transferring Transformer-based Image Compression from Human Perception to Machine Perception}.
\newblock \bibinfo{journal}{\emph{2023 IEEE/CVF International Conference on Computer Vision (ICCV)}} (\bibinfo{year}{2023}), \bibinfo{pages}{23240--23250}.
\newblock
\urldef\tempurl%
\url{https://api.semanticscholar.org/CorpusID:259108311}
\showURL{%
\tempurl}


\bibitem[Choi and Baji{\'c}(2021)]%
        {Choi2021LatentSpaceSF}
\bibfield{author}{\bibinfo{person}{Hyomin Choi} {and} \bibinfo{person}{Ivan~V. Baji{\'c}}.} \bibinfo{year}{2021}\natexlab{}.
\newblock \showarticletitle{Latent-Space Scalability for Multi-Task Collaborative Intelligence}.
\newblock \bibinfo{journal}{\emph{2021 IEEE International Conference on Image Processing (ICIP)}} (\bibinfo{year}{2021}), \bibinfo{pages}{3562--3566}.
\newblock
\urldef\tempurl%
\url{https://api.semanticscholar.org/CorpusID:235125754}
\showURL{%
\tempurl}


\bibitem[Choi and Bajić(2022)]%
        {scalableicm_choi}
\bibfield{author}{\bibinfo{person}{Hyomin Choi} {and} \bibinfo{person}{Ivan~V. Bajić}.} \bibinfo{year}{2022}\natexlab{}.
\newblock \showarticletitle{Scalable Image Coding for Humans and Machines}.
\newblock \bibinfo{journal}{\emph{IEEE Transactions on Image Processing}}  \bibinfo{volume}{31} (\bibinfo{year}{2022}), \bibinfo{pages}{2739--2754}.
\newblock
\urldef\tempurl%
\url{https://doi.org/10.1109/TIP.2022.3160602}
\showDOI{\tempurl}


\bibitem[Choi and Han(2020)]%
        {TAQNforJPEG}
\bibfield{author}{\bibinfo{person}{Jinyoung Choi} {and} \bibinfo{person}{Bohyung Han}.} \bibinfo{year}{2020}\natexlab{}.
\newblock \showarticletitle{Task-Aware Quantization Network for JPEG Image Compression}. In \bibinfo{booktitle}{\emph{Computer Vision -- ECCV 2020}}, \bibfield{editor}{\bibinfo{person}{Andrea Vedaldi}, \bibinfo{person}{Horst Bischof}, \bibinfo{person}{Thomas Brox}, {and} \bibinfo{person}{Jan-Michael Frahm}} (Eds.). \bibinfo{publisher}{Springer International Publishing}, \bibinfo{address}{Cham}, \bibinfo{pages}{309--324}.
\newblock
\showISBNx{978-3-030-58565-5}


\bibitem[Codevilla et~al\mbox{.}(2021)]%
        {Codevilla2021LearnedIC}
\bibfield{author}{\bibinfo{person}{Felipe Codevilla}, \bibinfo{person}{Jean-Gabriel Simard}, \bibinfo{person}{Ross Goroshin}, {and} \bibinfo{person}{Christopher~Joseph Pal}.} \bibinfo{year}{2021}\natexlab{}.
\newblock \showarticletitle{Learned Image Compression for Machine Perception}.
\newblock \bibinfo{journal}{\emph{ArXiv}}  \bibinfo{volume}{abs/2111.02249} (\bibinfo{year}{2021}).
\newblock
\urldef\tempurl%
\url{https://api.semanticscholar.org/CorpusID:241033392}
\showURL{%
\tempurl}


\bibitem[Cui et~al\mbox{.}(2024)]%
        {cui2024}
\bibfield{author}{\bibinfo{person}{Wenxue Cui}, \bibinfo{person}{Xingtao Wang}, \bibinfo{person}{Xiaopeng Fan}, \bibinfo{person}{Shaohui Liu}, \bibinfo{person}{Xinwei Gao}, {and} \bibinfo{person}{Debin Zhao}.} \bibinfo{year}{2024}\natexlab{}.
\newblock \showarticletitle{Deep Network for Image Compressed Sensing Coding Using Local Structural Sampling}.
\newblock \bibinfo{journal}{\emph{ACM Trans. Multimedia Comput. Commun. Appl.}} \bibinfo{volume}{20}, \bibinfo{number}{7}, Article \bibinfo{articleno}{212} (\bibinfo{date}{May} \bibinfo{year}{2024}), \bibinfo{numpages}{22}~pages.
\newblock
\showISSN{1551-6857}
\urldef\tempurl%
\url{https://doi.org/10.1145/3649441}
\showDOI{\tempurl}


\bibitem[Ding et~al\mbox{.}(2020)]%
        {ding2020local}
\bibfield{author}{\bibinfo{person}{Xintao Ding}, \bibinfo{person}{Qingde Li}, \bibinfo{person}{Yongqiang Cheng}, \bibinfo{person}{Jinbao Wang}, \bibinfo{person}{Weixin Bian}, {and} \bibinfo{person}{Biao Jie}.} \bibinfo{year}{2020}\natexlab{}.
\newblock \showarticletitle{Local keypoint-based Faster R-CNN}.
\newblock \bibinfo{journal}{\emph{Applied Intelligence}}  \bibinfo{volume}{50} (\bibinfo{year}{2020}), \bibinfo{pages}{3007--3022}.
\newblock


\bibitem[Duan et~al\mbox{.}(2020)]%
        {vcmaparadigm}
\bibfield{author}{\bibinfo{person}{Lingyu Duan}, \bibinfo{person}{Jiaying Liu}, \bibinfo{person}{Wenhan Yang}, \bibinfo{person}{Tiejun Huang}, {and} \bibinfo{person}{Wen Gao}.} \bibinfo{year}{2020}\natexlab{}.
\newblock \showarticletitle{Video Coding for Machines: A Paradigm of Collaborative Compression and Intelligent Analytics}.
\newblock \bibinfo{journal}{\emph{IEEE Transactions on Image Processing}}  \bibinfo{volume}{29} (\bibinfo{year}{2020}), \bibinfo{pages}{8680--8695}.
\newblock
\urldef\tempurl%
\url{https://doi.org/10.1109/TIP.2020.3016485}
\showDOI{\tempurl}


\bibitem[Duan et~al\mbox{.}(2016)]%
        {overviewcdvs}
\bibfield{author}{\bibinfo{person}{Ling-Yu Duan}, \bibinfo{person}{Vijay Chandrasekhar}, \bibinfo{person}{Jie Chen}, \bibinfo{person}{Jie Lin}, \bibinfo{person}{Zhe Wang}, \bibinfo{person}{Tiejun Huang}, \bibinfo{person}{Bernd Girod}, {and} \bibinfo{person}{Wen Gao}.} \bibinfo{year}{2016}\natexlab{}.
\newblock \showarticletitle{Overview of the MPEG-CDVS Standard}.
\newblock \bibinfo{journal}{\emph{Trans. Img. Proc.}} \bibinfo{volume}{25}, \bibinfo{number}{1}, \bibinfo{pages}{179–194}.
\newblock
\showISSN{1057-7149}
\urldef\tempurl%
\url{https://doi.org/10.1109/TIP.2015.2500034}
\showDOI{\tempurl}


\bibitem[Duan et~al\mbox{.}(2019)]%
        {cdvampeg}
\bibfield{author}{\bibinfo{person}{Ling-Yu Duan}, \bibinfo{person}{Yihang Lou}, \bibinfo{person}{Yan Bai}, \bibinfo{person}{Tiejun Huang}, \bibinfo{person}{Wen Gao}, \bibinfo{person}{Vijay Chandrasekhar}, \bibinfo{person}{Jie Lin}, \bibinfo{person}{Shiqi Wang}, {and} \bibinfo{person}{Alex~Chichung Kot}.} \bibinfo{year}{2019}\natexlab{}.
\newblock \showarticletitle{Compact Descriptors for Video Analysis: The Emerging MPEG Standard}.
\newblock \bibinfo{journal}{\emph{IEEE MultiMedia}} \bibinfo{volume}{26}, \bibinfo{number}{2} (\bibinfo{year}{2019}), \bibinfo{pages}{44--54}.
\newblock
\urldef\tempurl%
\url{https://doi.org/10.1109/MMUL.2018.2873844}
\showDOI{\tempurl}


\bibitem[Feng et~al\mbox{.}(2022)]%
        {Omni-ICM}
\bibfield{author}{\bibinfo{person}{Ruoyu Feng}, \bibinfo{person}{Xin Jin}, \bibinfo{person}{Zongyu Guo}, \bibinfo{person}{Runsen Feng}, \bibinfo{person}{Yixin Gao}, \bibinfo{person}{Tianyu He}, \bibinfo{person}{Zhizheng Zhang}, \bibinfo{person}{Simeng Sun}, {and} \bibinfo{person}{Zhibo Chen}.} \bibinfo{year}{2022}\natexlab{}.
\newblock \showarticletitle{Image Coding for Machines with Omnipotent Feature Learning}. In \bibinfo{booktitle}{\emph{Computer Vision -- ECCV 2022}}, \bibfield{editor}{\bibinfo{person}{Shai Avidan}, \bibinfo{person}{Gabriel Brostow}, \bibinfo{person}{Moustapha Ciss{\'e}}, \bibinfo{person}{Giovanni~Maria Farinella}, {and} \bibinfo{person}{Tal Hassner}} (Eds.). \bibinfo{publisher}{Springer Nature Switzerland}, \bibinfo{address}{Cham}, \bibinfo{pages}{510--528}.
\newblock
\showISBNx{978-3-031-19836-6}


\bibitem[Gao et~al\mbox{.}(2021)]%
        {Gao2021RecentSD}
\bibfield{author}{\bibinfo{person}{Wen Gao}, \bibinfo{person}{Shan Liu}, \bibinfo{person}{Xiaozhong Xu}, \bibinfo{person}{Manouchehr Rafie}, \bibinfo{person}{Yuan Zhang}, {and} \bibinfo{person}{Igor D.~D. Curcio}.} \bibinfo{year}{2021}\natexlab{}.
\newblock \showarticletitle{Recent Standard Development Activities on Video Coding for Machines}.
\newblock \bibinfo{journal}{\emph{ArXiv}}  \bibinfo{volume}{abs/2105.12653} (\bibinfo{year}{2021}).
\newblock
\urldef\tempurl%
\url{https://api.semanticscholar.org/CorpusID:235195868}
\showURL{%
\tempurl}


\bibitem[He et~al\mbox{.}(2022)]%
        {He2022ELICEL}
\bibfield{author}{\bibinfo{person}{Dailan He}, \bibinfo{person}{Zi Yang}, \bibinfo{person}{Weikun Peng}, \bibinfo{person}{Rui Ma}, \bibinfo{person}{Hongwei Qin}, {and} \bibinfo{person}{Yan Wang}.} \bibinfo{year}{2022}\natexlab{}.
\newblock \showarticletitle{ELIC: Efficient Learned Image Compression with Unevenly Grouped Space-Channel Contextual Adaptive Coding}.
\newblock \bibinfo{journal}{\emph{2022 IEEE/CVF Conference on Computer Vision and Pattern Recognition (CVPR)}} (\bibinfo{year}{2022}), \bibinfo{pages}{5708--5717}.
\newblock
\urldef\tempurl%
\url{https://api.semanticscholar.org/CorpusID:247594672}
\showURL{%
\tempurl}


\bibitem[He et~al\mbox{.}(2017)]%
        {maskrcnn}
\bibfield{author}{\bibinfo{person}{Kaiming He}, \bibinfo{person}{Georgia Gkioxari}, \bibinfo{person}{Piotr Dollár}, {and} \bibinfo{person}{Ross Girshick}.} \bibinfo{year}{2017}\natexlab{}.
\newblock \showarticletitle{Mask R-CNN}. In \bibinfo{booktitle}{\emph{2017 IEEE International Conference on Computer Vision (ICCV)}}. \bibinfo{publisher}{IEEE}, \bibinfo{address}{Venice, Italy}, \bibinfo{pages}{2980--2988}.
\newblock
\urldef\tempurl%
\url{https://doi.org/10.1109/ICCV.2017.322}
\showDOI{\tempurl}


\bibitem[He et~al\mbox{.}(2016)]%
        {7780459}
\bibfield{author}{\bibinfo{person}{Kaiming He}, \bibinfo{person}{Xiangyu Zhang}, \bibinfo{person}{Shaoqing Ren}, {and} \bibinfo{person}{Jian Sun}.} \bibinfo{year}{2016}\natexlab{}.
\newblock \showarticletitle{Deep Residual Learning for Image Recognition}. In \bibinfo{booktitle}{\emph{2016 IEEE Conference on Computer Vision and Pattern Recognition (CVPR)}}. \bibinfo{publisher}{IEEE}, \bibinfo{address}{Las Vegas, NV, USA}, \bibinfo{pages}{770--778}.
\newblock
\urldef\tempurl%
\url{https://doi.org/10.1109/CVPR.2016.90}
\showDOI{\tempurl}


\bibitem[Hu et~al\mbox{.}(2020)]%
        {hu2020sensitivity}
\bibfield{author}{\bibinfo{person}{Yuzhang Hu}, \bibinfo{person}{Sifeng Xia}, \bibinfo{person}{Wenhan Yang}, {and} \bibinfo{person}{Jiaying Liu}.} \bibinfo{year}{2020}\natexlab{}.
\newblock \showarticletitle{Sensitivity-aware bit allocation for intermediate deep feature compression}. In \bibinfo{booktitle}{\emph{2020 IEEE International Conference on Visual Communications and Image Processing (VCIP)}}. IEEE, \bibinfo{publisher}{IEEE}, \bibinfo{address}{Macau, China}, \bibinfo{pages}{475--478}.
\newblock


\bibitem[Huang et~al\mbox{.}(2021)]%
        {analysisRD2021}
\bibfield{author}{\bibinfo{person}{Zhimeng Huang}, \bibinfo{person}{Chuanmin Jia}, \bibinfo{person}{Shanshe Wang}, {and} \bibinfo{person}{Siwei Ma}.} \bibinfo{year}{2021}\natexlab{}.
\newblock \showarticletitle{Visual Analysis Motivated Rate-Distortion Model for Image Coding}. In \bibinfo{booktitle}{\emph{2021 IEEE International Conference on Multimedia and Expo (ICME)}}. \bibinfo{publisher}{IEEE}, \bibinfo{address}{Shenzhen, China}, \bibinfo{pages}{1--6}.
\newblock
\urldef\tempurl%
\url{https://doi.org/10.1109/ICME51207.2021.9428417}
\showDOI{\tempurl}


\bibitem[Le et~al\mbox{.}(2021)]%
        {ICMcontent2021}
\bibfield{author}{\bibinfo{person}{Nam Le}, \bibinfo{person}{Honglei Zhang}, \bibinfo{person}{Francesco Cricri}, \bibinfo{person}{Ramin Ghaznavi-Youvalari}, \bibinfo{person}{Hamed~Rezazadegan Tavakoli}, {and} \bibinfo{person}{Esa Rahtu}.} \bibinfo{year}{2021}\natexlab{}.
\newblock \showarticletitle{Learned Image Coding for Machines: A Content-Adaptive Approach}. In \bibinfo{booktitle}{\emph{2021 IEEE International Conference on Multimedia and Expo (ICME)}}. \bibinfo{publisher}{IEEE}, \bibinfo{address}{Shenzhen, China}, \bibinfo{pages}{1--6}.
\newblock
\urldef\tempurl%
\url{https://doi.org/10.1109/ICME51207.2021.9428224}
\showDOI{\tempurl}


\bibitem[Lin et~al\mbox{.}(2018)]%
        {lin2018focal}
\bibfield{author}{\bibinfo{person}{Tsung-Yi Lin}, \bibinfo{person}{Priya Goyal}, \bibinfo{person}{Ross Girshick}, \bibinfo{person}{Kaiming He}, {and} \bibinfo{person}{Piotr Dollár}.} \bibinfo{year}{2018}\natexlab{}.
\newblock \bibinfo{title}{Focal Loss for Dense Object Detection}.
\newblock
\newblock
\showeprint[arxiv]{1708.02002}


\bibitem[Liu et~al\mbox{.}(2022)]%
        {improvingmvtincompressdomain}
\bibfield{author}{\bibinfo{person}{Jinming Liu}, \bibinfo{person}{Heming Sun}, {and} \bibinfo{person}{Jiro Katto}.} \bibinfo{year}{2022}\natexlab{}.
\newblock \showarticletitle{Improving Multiple Machine Vision Tasks in the Compressed Domain}. In \bibinfo{booktitle}{\emph{2022 26th International Conference on Pattern Recognition (ICPR)}}. \bibinfo{publisher}{IEEE}, \bibinfo{address}{Montreal, QC, Canada}, \bibinfo{pages}{331--337}.
\newblock
\urldef\tempurl%
\url{https://doi.org/10.1109/ICPR56361.2022.9956532}
\showDOI{\tempurl}


\bibitem[Liu et~al\mbox{.}(2023a)]%
        {LICTCM}
\bibfield{author}{\bibinfo{person}{Jinming Liu}, \bibinfo{person}{Heming Sun}, {and} \bibinfo{person}{Jiro Katto}.} \bibinfo{year}{2023}\natexlab{a}.
\newblock \showarticletitle{Learned Image Compression with Mixed Transformer-CNN Architectures}. In \bibinfo{booktitle}{\emph{2023 IEEE/CVF Conference on Computer Vision and Pattern Recognition (CVPR)}}. \bibinfo{publisher}{IEEE}, \bibinfo{address}{Vancouver, BC, Canada}, \bibinfo{pages}{14388--14397}.
\newblock
\urldef\tempurl%
\url{https://doi.org/10.1109/CVPR52729.2023.01383}
\showDOI{\tempurl}


\bibitem[Liu et~al\mbox{.}(2021)]%
        {Liu2021SemanticstoSignalSI}
\bibfield{author}{\bibinfo{person}{Kang Liu}, \bibinfo{person}{Dong Liu}, \bibinfo{person}{Li Li}, \bibinfo{person}{Ning Yan}, {and} \bibinfo{person}{Houqiang Li}.} \bibinfo{year}{2021}\natexlab{}.
\newblock \showarticletitle{Semantics-to-Signal Scalable Image Compression with Learned Revertible Representations}.
\newblock \bibinfo{journal}{\emph{International Journal of Computer Vision}}  \bibinfo{volume}{129} (\bibinfo{year}{2021}), \bibinfo{pages}{2605 -- 2621}.
\newblock
\urldef\tempurl%
\url{https://api.semanticscholar.org/CorpusID:236930362}
\showURL{%
\tempurl}


\bibitem[Liu et~al\mbox{.}(2023b)]%
        {mutualFCMV2023}
\bibfield{author}{\bibinfo{person}{Tie Liu}, \bibinfo{person}{Mai Xu}, \bibinfo{person}{Shengxi Li}, \bibinfo{person}{Chaoran Chen}, \bibinfo{person}{Li Yang}, {and} \bibinfo{person}{Zhuoyi Lv}.} \bibinfo{year}{2023}\natexlab{b}.
\newblock \showarticletitle{Learnt Mutual Feature Compression for Machine Vision}. In \bibinfo{booktitle}{\emph{ICASSP 2023 - 2023 IEEE International Conference on Acoustics, Speech and Signal Processing (ICASSP)}}. \bibinfo{publisher}{IEEE}, \bibinfo{address}{Rhodes Island, Greece}, \bibinfo{pages}{1--5}.
\newblock
\urldef\tempurl%
\url{https://doi.org/10.1109/ICASSP49357.2023.10094830}
\showDOI{\tempurl}


\bibitem[Peng and Hang(2020)]%
        {vcm}
\bibfield{author}{\bibinfo{person}{Wen-Hsiao Peng} {and} \bibinfo{person}{Hsueh-Ming Hang}.} \bibinfo{year}{2020}\natexlab{}.
\newblock \showarticletitle{Recent Advances in End-to-End Learned Image and Video Compression}. In \bibinfo{booktitle}{\emph{2020 IEEE International Conference on Visual Communications and Image Processing (VCIP)}}. \bibinfo{publisher}{IEEE}, \bibinfo{address}{Macau, China}, \bibinfo{pages}{1--2}.
\newblock
\urldef\tempurl%
\url{https://doi.org/10.1109/VCIP49819.2020.9301753}
\showDOI{\tempurl}


\bibitem[Ren et~al\mbox{.}(2017)]%
        {fasterrcnn}
\bibfield{author}{\bibinfo{person}{Shaoqing Ren}, \bibinfo{person}{Kaiming He}, \bibinfo{person}{Ross Girshick}, {and} \bibinfo{person}{Jian Sun}.} \bibinfo{year}{2017}\natexlab{}.
\newblock \showarticletitle{Faster R-CNN: Towards Real-Time Object Detection with Region Proposal Networks}.
\newblock \bibinfo{journal}{\emph{IEEE Transactions on Pattern Analysis and Machine Intelligence}} \bibinfo{volume}{39}, \bibinfo{number}{6} (\bibinfo{year}{2017}), \bibinfo{pages}{1137--1149}.
\newblock
\urldef\tempurl%
\url{https://doi.org/10.1109/TPAMI.2016.2577031}
\showDOI{\tempurl}


\bibitem[Shindo et~al\mbox{.}(2024)]%
        {shindo2024image}
\bibfield{author}{\bibinfo{person}{Takahiro Shindo}, \bibinfo{person}{Kein Yamada}, \bibinfo{person}{Taiju Watanabe}, {and} \bibinfo{person}{Hiroshi Watanabe}.} \bibinfo{year}{2024}\natexlab{}.
\newblock \showarticletitle{Image Coding for Machines with Edge Information Learning Using Segment Anything}.
\newblock \bibinfo{journal}{\emph{arXiv preprint arXiv:2403.04173}} (\bibinfo{year}{2024}).
\newblock


\bibitem[Tu et~al\mbox{.}(2024)]%
        {Tu2024}
\bibfield{author}{\bibinfo{person}{Hanyue Tu}, \bibinfo{person}{Li Li}, \bibinfo{person}{Wengang Zhou}, {and} \bibinfo{person}{Houqiang Li}.} \bibinfo{year}{2024}\natexlab{}.
\newblock \showarticletitle{Reconstruction-free Image Compression for Machine Vision via Knowledge Transfer}.
\newblock \bibinfo{journal}{\emph{ACM Trans. Multimedia Comput. Commun. Appl.}} (\bibinfo{date}{July} \bibinfo{year}{2024}).
\newblock
\showISSN{1551-6857}
\urldef\tempurl%
\url{https://doi.org/10.1145/3678471}
\showDOI{\tempurl}
\newblock
\shownote{Just Accepted}.


\bibitem[Wang et~al\mbox{.}(2021b)]%
        {multigranularity2021wangshurun}
\bibfield{author}{\bibinfo{person}{Shurun Wang}, \bibinfo{person}{Shiqi Wang}, \bibinfo{person}{Wenhan Yang}, \bibinfo{person}{Xinfeng Zhang}, \bibinfo{person}{Shanshe Wang}, {and} \bibinfo{person}{Siwei Ma}.} \bibinfo{year}{2021}\natexlab{b}.
\newblock \showarticletitle{Teacher-Student Learning With Multi-Granularity Constraint Towards Compact Facial Feature Representation}. In \bibinfo{booktitle}{\emph{ICASSP 2021 - 2021 IEEE International Conference on Acoustics, Speech and Signal Processing (ICASSP)}}. \bibinfo{publisher}{Institute of Electrical and Electronics Engineers, Inc.}, \bibinfo{address}{United States}, \bibinfo{pages}{8503--8507}.
\newblock
\urldef\tempurl%
\url{https://doi.org/10.1109/ICASSP39728.2021.9413506}
\showDOI{\tempurl}


\bibitem[Wang et~al\mbox{.}(2021a)]%
        {endtoend2021wang}
\bibfield{author}{\bibinfo{person}{Shurun Wang}, \bibinfo{person}{Zhao Wang}, \bibinfo{person}{Shiqi Wang}, {and} \bibinfo{person}{Yan Ye}.} \bibinfo{year}{2021}\natexlab{a}.
\newblock \showarticletitle{End-to-End Compression Towards Machine Vision: Network Architecture Design and Optimization}.
\newblock \bibinfo{journal}{\emph{IEEE Open Journal of Circuits and Systems}}  \bibinfo{volume}{2} (\bibinfo{year}{2021}), \bibinfo{pages}{675--685}.
\newblock
\urldef\tempurl%
\url{https://doi.org/10.1109/OJCAS.2021.3126061}
\showDOI{\tempurl}


\bibitem[Wang et~al\mbox{.}(2023)]%
        {deepicm2023wang}
\bibfield{author}{\bibinfo{person}{Shurun Wang}, \bibinfo{person}{Zhao Wang}, \bibinfo{person}{Shiqi Wang}, {and} \bibinfo{person}{Yan Ye}.} \bibinfo{year}{2023}\natexlab{}.
\newblock \showarticletitle{Deep Image Compression Toward Machine Vision: A Unified Optimization Framework}.
\newblock \bibinfo{journal}{\emph{IEEE Transactions on Circuits and Systems for Video Technology}} \bibinfo{volume}{33}, \bibinfo{number}{6} (\bibinfo{year}{2023}), \bibinfo{pages}{2979--2989}.
\newblock
\urldef\tempurl%
\url{https://doi.org/10.1109/TCSVT.2022.3230843}
\showDOI{\tempurl}


\bibitem[Wang et~al\mbox{.}(2022)]%
        {interactionorientedicm_iot}
\bibfield{author}{\bibinfo{person}{Zixi Wang}, \bibinfo{person}{Fan Li}, \bibinfo{person}{Jing Xu}, {and} \bibinfo{person}{Pamela~C. Cosman}.} \bibinfo{year}{2022}\natexlab{}.
\newblock \showarticletitle{Human–Machine Interaction-Oriented Image Coding for Resource-Constrained Visual Monitoring in IoT}.
\newblock \bibinfo{journal}{\emph{IEEE Internet of Things Journal}} \bibinfo{volume}{9}, \bibinfo{number}{17} (\bibinfo{year}{2022}), \bibinfo{pages}{16181--16195}.
\newblock
\urldef\tempurl%
\url{https://doi.org/10.1109/JIOT.2022.3150417}
\showDOI{\tempurl}


\bibitem[Yan et~al\mbox{.}(2020)]%
        {SSICMfeature}
\bibfield{author}{\bibinfo{person}{Ning Yan}, \bibinfo{person}{Dong Liu}, \bibinfo{person}{Houqiang Li}, {and} \bibinfo{person}{Feng Wu}.} \bibinfo{year}{2020}\natexlab{}.
\newblock \showarticletitle{Semantically Scalable Image Coding With Compression of Feature Maps}. In \bibinfo{booktitle}{\emph{2020 IEEE International Conference on Image Processing (ICIP)}}. \bibinfo{publisher}{IEEE}, \bibinfo{address}{Abu Dhabi, United Arab Emirates}, \bibinfo{pages}{3114--3118}.
\newblock
\urldef\tempurl%
\url{https://doi.org/10.1109/ICIP40778.2020.9191184}
\showDOI{\tempurl}


\bibitem[Yang et~al\mbox{.}(2021)]%
        {towardsSIC2021}
\bibfield{author}{\bibinfo{person}{Shuai Yang}, \bibinfo{person}{Yueyu Hu}, \bibinfo{person}{Wenhan Yang}, \bibinfo{person}{Ling-Yu Duan}, {and} \bibinfo{person}{Jiaying Liu}.} \bibinfo{year}{2021}\natexlab{}.
\newblock \showarticletitle{Towards Coding for Human and Machine Vision: Scalable Face Image Coding}.
\newblock \bibinfo{journal}{\emph{IEEE Transactions on Multimedia}}  \bibinfo{volume}{23} (\bibinfo{year}{2021}), \bibinfo{pages}{2957--2971}.
\newblock
\urldef\tempurl%
\url{https://doi.org/10.1109/TMM.2021.3068580}
\showDOI{\tempurl}


\bibitem[Yang et~al\mbox{.}(2024)]%
        {compactVCM2024}
\bibfield{author}{\bibinfo{person}{Wenhan Yang}, \bibinfo{person}{Haofeng Huang}, \bibinfo{person}{Yueyu Hu}, \bibinfo{person}{Ling-Yu Duan}, {and} \bibinfo{person}{Jiaying Liu}.} \bibinfo{year}{2024}\natexlab{}.
\newblock \showarticletitle{Video Coding for Machines: Compact Visual Representation Compression for Intelligent Collaborative Analytics}.
\newblock \bibinfo{journal}{\emph{IEEE Transactions on Pattern Analysis and Machine Intelligence}} \bibinfo{volume}{46}, \bibinfo{number}{7} (\bibinfo{year}{2024}), \bibinfo{pages}{5174--5191}.
\newblock
\urldef\tempurl%
\url{https://doi.org/10.1109/TPAMI.2024.3367293}
\showDOI{\tempurl}


\bibitem[Yuan et~al\mbox{.}(2015)]%
        {RDO2015}
\bibfield{author}{\bibinfo{person}{Hui Yuan}, \bibinfo{person}{Sam Kwong}, \bibinfo{person}{Xu Wang}, \bibinfo{person}{Wei Gao}, {and} \bibinfo{person}{Yun Zhang}.} \bibinfo{year}{2015}\natexlab{}.
\newblock \showarticletitle{Rate Distortion Optimized Inter-View Frame Level Bit Allocation Method for MV-HEVC}.
\newblock \bibinfo{journal}{\emph{IEEE Transactions on Multimedia}} \bibinfo{volume}{17}, \bibinfo{number}{12} (\bibinfo{year}{2015}), \bibinfo{pages}{2134--2146}.
\newblock
\urldef\tempurl%
\url{https://doi.org/10.1109/TMM.2015.2477682}
\showDOI{\tempurl}


\bibitem[Zhang et~al\mbox{.}(2024b)]%
        {zhang2024}
\bibfield{author}{\bibinfo{person}{Gai Zhang}, \bibinfo{person}{Xinfeng Zhang}, {and} \bibinfo{person}{Lv Tang}.} \bibinfo{year}{2024}\natexlab{b}.
\newblock \showarticletitle{Unified and Scalable Deep Image Compression Framework for Human and Machine}.
\newblock \bibinfo{journal}{\emph{ACM Trans. Multimedia Comput. Commun. Appl.}} (\bibinfo{date}{July} \bibinfo{year}{2024}).
\newblock
\showISSN{1551-6857}
\urldef\tempurl%
\url{https://doi.org/10.1145/3678472}
\showDOI{\tempurl}
\newblock
\shownote{Just Accepted}.


\bibitem[Zhang et~al\mbox{.}(2023)]%
        {zhang2023rethinking}
\bibfield{author}{\bibinfo{person}{Pingping Zhang}, \bibinfo{person}{Shiqi Wang}, \bibinfo{person}{Meng Wang}, \bibinfo{person}{Jiguo Li}, \bibinfo{person}{Xu Wang}, {and} \bibinfo{person}{Sam Kwong}.} \bibinfo{year}{2023}\natexlab{}.
\newblock \showarticletitle{Rethinking semantic image compression: Scalable representation with cross-modality transfer}.
\newblock \bibinfo{journal}{\emph{IEEE Transactions on circuits and systems for video technology}} \bibinfo{volume}{33}, \bibinfo{number}{8} (\bibinfo{year}{2023}), \bibinfo{pages}{4441--4445}.
\newblock


\bibitem[Zhang et~al\mbox{.}(2024a)]%
        {zhangyun2024}
\bibfield{author}{\bibinfo{person}{Yun Zhang}, \bibinfo{person}{Haoqin Lin}, \bibinfo{person}{Jing Sun}, \bibinfo{person}{Linwei Zhu}, {and} \bibinfo{person}{Sam Kwong}.} \bibinfo{year}{2024}\natexlab{a}.
\newblock \showarticletitle{Learning to Predict Object-Wise Just Recognizable Distortion for Image and Video Compression}.
\newblock \bibinfo{journal}{\emph{IEEE Transactions on Multimedia}}  \bibinfo{volume}{26} (\bibinfo{year}{2024}), \bibinfo{pages}{5925--5938}.
\newblock
\urldef\tempurl%
\url{https://doi.org/10.1109/TMM.2023.3340882}
\showDOI{\tempurl}


\bibitem[Özyılkan et~al\mbox{.}(2023)]%
        {latentSICM}
\bibfield{author}{\bibinfo{person}{Ezgi Özyılkan}, \bibinfo{person}{Mateen Ulhaq}, \bibinfo{person}{Hyomin Choi}, {and} \bibinfo{person}{Fabien Racapé}.} \bibinfo{year}{2023}\natexlab{}.
\newblock \showarticletitle{Learned Disentangled Latent Representations for Scalable Image Coding for Humans and Machines}. In \bibinfo{booktitle}{\emph{2023 Data Compression Conference (DCC)}}. \bibinfo{publisher}{IEEE}, \bibinfo{address}{Snowbird, UT, USA}, \bibinfo{pages}{42--51}.
\newblock
\urldef\tempurl%
\url{https://doi.org/10.1109/DCC55655.2023.00012}
\showDOI{\tempurl}


\end{thebibliography}

\end{document}